\documentclass[10pt]{article}

\usepackage{helvet} 
\usepackage{paratype} 
\usepackage{amsmath}
\usepackage{textcase}
\usepackage{fullpage, setspace, parskip, titlesec}
\usepackage[section]{placeins}
\usepackage[dvipsnames]{xcolor}
\usepackage{breakcites, lineno, hyphenat, rotating, etoolbox}
\usepackage{ graphicx}
\usepackage{latexsym, textcomp, longtable, tabulary, tabularx}
\usepackage{booktabs,array,multirow}
\usepackage{amsfonts,amsmath,amssymb}
\usepackage[utf8]{inputenc}
\usepackage[english]{babel}
\usepackage{xspace, caption}
\usepackage[toc,symbols]{glossaries}
\usepackage{bm}
\usepackage[automake]{glossaries-extra}
\usepackage[round]{natbib}
\usepackage[left=0.75in]{geometry}
\usepackage[ampersand]{easylist}
\usepackage{enumitem}
\usepackage{pifont, makecell}
\usepackage[disable]{todonotes}
\usepackage{float}
\usepackage{lscape}
\usepackage{tcolorbox}
\usepackage{titlesec}
\usepackage{latexcolors}
\tcbuselibrary{most}
\usepackage{epigraph}
\usepackage{svg}
\usepackage{multicol}
\PassOptionsToPackage{hyphens}{url}
\usepackage[colorlinks]{hyperref}
\usepackage{url}

\newcounter{panel}
\newenvironment{panel}[1][]{\refstepcounter{panel}\par\medskip
   \textbf{Panel~\thepanel. #1} \rmfamily}{\medskip}

\definecolor{darkblue}{RGB}{0,0,130}
\definecolor{darkgreen}{RGB}{0,80,20}
\hypersetup{colorlinks = true, linkcolor = darkblue, citecolor = darkgreen, urlcolor=	RedViolet}

\glsenablehyper
\newcommand{\orcid}[1]{\href{https://orcid.org/#1}{}}

\AtBeginDocument{\DeclareGraphicsExtensions{.png,.PNG,.jpg,.JPG,.jpeg,.JPEG}}

\renewenvironment{abstract}
 {{\noindent\bfseries{\abstractname}\par\nobreak}\footnotesize}
 {\bigskip}

\titlespacing{\section}{0pt}{*3}{*1}
\titlespacing{\subsection}{0pt}{*2}{*0.5}
\titlespacing{\subsubsection}{0pt}{*1.5}{0pt}
\providecommand\citet{\cite}
\providecommand\citep{\cite}

\renewcommand\cite{\citep}

\newglossary[odg]{oalgo}{old}{odn}{OHD-GAN Acronyms}
\setabbreviationstyle[acronym]{long-short}
\glsxtraddallcrossrefs
\makeglossaries
\newglossaryentry{t-sne}{
    type=\acronymtype,
    name={t-SNE}, 
    description={The t-Distributed Stochastic Neighbor Embedding clustering algorithm is a nonlinear dimensionality reduction technique commonly applied to high-dimensional data. See \citet{maaten2008tsne}.}
    text={t-SNE},
    first={t-Distributed Stochastic Neighbor Embedding (t-SNE)}
}

\newglossaryentry{ward2icu}{
    name={Ward2ICU},
    description={Labelled synthetic dataset of vital signs indicating patient transitions from the ward to intensive care units generated by \citeauthor{severo2019ward2icu} using a \gls{cwgan-gp}.}
}

\newglossaryentry{mode-collapse}{
    type=\acronymtype,
    name={mode collapse}, 
    description={Mode collapse occurs when the \gls{gan} training procedure fails to converge, or converges to an undesirable local minima resulting in a lack of variety in the generated samples.}
    text={mode collapse},
    first={mode collapse}
}

\newglossaryentry{mb-avg}{
    type=\acronymtype,
    name={MB-Avg}, 
    description={Alternative to mini-batch discrimination that performs well on categorical features to cope with mode collapse, see \cite{Choi2017-nt}}
    text={MB-Avg},
    first={Mini-batch Averaging (MB-Avg)}
}

\newglossaryentry{dom-tran}{
    type=\acronymtype,
    name={DT}, 
    description={Applying a function to data points transforming them from one domain or category to another.}
    text={DT},
    first={Domain Translation (DT)}
}

\newglossaryentry{semi-sup}{
    type=\acronymtype,
    name={SSL}, 
    description={Semi-supervised learning refers to a type of \gls{ml} algorithm training procedure. Where in supervised learning all the data points are labelled and the algorithm is trained conditionally, and in unsupervised learning the data is unlabelled leaving the algorithm to discover patterns in the data, in semi-supervised learning only a small portion of the data is labelled. There is multitude of gls{ssl} algorithms and application, which generally involve learning from the labelled data points to gather information from the unlabelled. \cite{ssl-mit} }
    text={SSL},
    first={Semi-supervised learning (SSL)}
}
\newglossaryentry{re-iden}{
    type=\acronymtype,
    name={re-identification attack}, 
    description={See \gls{mi}}
    text={re-identification attack},
    first={re-identification attack}
}
\newglossaryentry{dbio}{
    type=\acronymtype,
    name={digital bio-markers}, 
    description={As opposed to classical bio-markers, which are broadly defined as any chemical, physical or biological indication of a patient's state the can be measured and are reproducible \cite{biomarkers2010}. Digital bio-markers are a trend emerging from the ubiquity of personal electronics devices which are said to have great potential as equivalent indicators, described as "[...] objective, quantifiable, physiological, and behavioural measures that are collected by sensors embedded in portable, wearable, implantable, or digestible devices." \cite{digital2020}
    text={digital bio-markers},
    first={digital bio-markers}
}}

\newglossaryentry{exploding}{
    type=\acronymtype,
    name={exploding gradients}, 
    description={When training a GAN, the gradients can accumulate large amounts of error, destabilising or disabling the training procedure.}
    text={exploding gradients},
    first={exploding gradients}
}

\newglossaryentry{vanishing}{
    type=\acronymtype,
    name={vanishing gradients}, 
    description={When training a GAN, the gradients become null and the network can no longer be updated.}
    text={vanishing gradients},
    first={vanishing gradients}
}

\newglossaryentry{a-disclosure}{
    type=\acronymtype,
    name={AD}, 
    description={Occures when an attacker can uncover confidential attributes of and individual from those released in an anonymized form.}
    text={AD},
    first={Attribute disclosure}
}

\newglossaryentry{mar}{
    type=\acronymtype,
    name={MaR}, 
    description={Given a dataset with missing entries , the missingness depends only on the observed variables \cite{yoon2018imputation}.}
    text={MaR},
    first={Missing at Random (MaR)}
}

\newglossaryentry{mcar}{
    type=\acronymtype,
    name={MCaR}, 
    description={Missing Completely at Random, description={Given a dataset with missing entries, the missingness is not dependant on any of the variables, thus occurs completely at random \cite{yoon2018imputation}.}}
    text={MCaR},
    first={Missing Completly at Random (MCaR)}
}

\newglossaryentry{}{
    type=\acronymtype,
    name={}, 
    description={}
    text={},
    first={}
}

\newglossaryentry{msn}{
 type=\acronymtype,
 name={MSN},
 description={Per feature, a variational Gaussian mixture model is used to estimate the number of modes and fit a Gaussian mixture. A one-hot vector indicating the mode, and a scalar indicating the value within the mode is produced. See \cite{Xu2019-ay}.},
 text={MSN},
 first={Mode-specific normalization (MSN)}
}

\newglossaryentry{tbs}{
    type=\acronymtype,
    name={TbS}, 
    description={To deal with the imbalance of values in categorical featues, during training the data is resampled in a way that all the categories from discrete attributes are sampled evenly, without inducing bias and so as to recover real data distribution. See \cite{Xu2019-ay} for a step-by-step spefication.}
    text={TbS},
    first={Training by sampling (TbS)}  
}

\newglossaryentry{nnaa}{
    type=\acronymtype,
    name={NN-AA}, 
    description={"Compares the distance from one point in a target distribution T, to the nearest point in a source distribution S, to the distance to the next nearest point in the target distribution." See \cite{yale2019ESANN}.}
    text={NN-AA},
    first={Nearest-neighbor Adversarial Accuracy (NN-AA)}
}

\newglossaryentry{pl}{
    type=\acronymtype,
    name={PL}, 
    description={Difference of \gls{nnaa} on the test set and on the training set. See \cite{yale2019ESANN}.}
    text={PL},
    first={Privacy loss (PL)}
}

\newglossaryentry{dt}{
    type=\acronymtype,
    name={DT}, 
    description={The discriminator is tested on batches of synthetic data produced by other methods to asses the possibility of over-fitting, see \cite{yale2019ESANN}.}
    text={DT},
    first={Discriminator testing (DT)}
}

\newglossaryentry{eicu}{
    type=\acronymtype,
    name={eICU}, 
    description={}
    text={eICU},
    first={Telehealth for the Intensive Care Unit (eICU)}
}

\newglossaryentry{do}{
    type=\acronymtype,
    name={DO}, 
    description={Privacy preservation method. See \cite{yale2019ESANN} based on \cite{Dwork2008, Prasser2017}.}
    text={DO},
    first={Data obfuscation (DO}
}

\newglossaryentry{pate}{
    type=\acronymtype,
    name={PATE}, 
    description={Differential privacy method, best described by \citeauthor{Papernot2017}: "The approach combines, in a black-box fashion, multiple models trained with disjoint datasets, such as records from different subsets of users. Because they rely directly on sensitive data, these models are not published, but instead used as "teachers" for a "student" model. The student learns to predict an output chosen by noisy voting among all of the teachers, and cannot directly access an individual teacher or the underlying data or parameters. The student's privacy properties can be understood both intuitively (since no single teacher and thus no single dataset dictates the student's training) and formally, in terms of differential privacy." \cite{Papernot2017,Papernot2018}}
    text={PATE},
    first={Private Aggregation of Teacher Ensembles (PATE)}
}

\newglossaryentry{mbd}{
    type=\acronymtype,
    name={MBD}, 
    description={Training technique. See \cite{Salimans2016}}
    text={MBD},
    first={Mini-batch discrimination (MDB)}
}

\newglossaryentry{t-gan}{
    type=\acronymtype,
    name={T-GAN}, 
    description={Training technique to stabilise training. Allows the introduction of real sample information into the process of training the the generator. See \cite{Jolicoeur-Martineau2019, Su2018}}
    text={T-GAN},
    first={Turing \gls{gan}}
}

\newglossaryentry{corrnn}{
    type=\acronymtype,
    name={CorrNN}, 
    description={Learns a common representation of two views, taking into account their correlation. See \cite{Jolicoeur-Martineau2019, Su2018}}
    text={CorrNN},
    first={Correlation \gls{nn}}
}

\newglossaryentry{fullbba}{
    type=\acronymtype,
    name={FullBB}, 
    description={A \gls{mi} attack setting where an attacker has now knowledge of the internal workings of the generator, but can only sample from it.}
    text={FullBB},
    first={Full Black-box Attack}
}

\newglossaryentry{partbba}{
    type=\acronymtype,
    name={PartBB}, 
    description={Similar to to the \gls{fullbba} setting with the attacker having the additional knowledge about the latent input $z$.}
    text={PartBB},
    first={Partial Black-box Attack}
}

\newglossaryentry{wba}{
    type=\acronymtype,
    name={WBA},
    description={Similar to the \gls{partbba} and \gls{fullbba} settings, but with the attacker having full knowledge of the generator internals, including gradient information.},
    text={WBA},
    first={White-box Attack}
}

\newglossaryentry{bgan}{
    type=\acronymtype,
    name={BGAN},
    description={Method for training GANs with discrete data that uses the estimated difference measure from the discriminator to compute importance weights for generated samples, enabling back-propagation. Tends to push the generated samples to lie on the decision boundary of the discriminator, which also improves stability of training on continuous data \cite{hjelm2017boundaryseeking}},
    text={BGAN},
    first={Boundary-seeking \gls{gan}}
}

\newglossaryentry{sdv}{
    type=\acronymtype,
    name={SDV},
    description={Generative model for relational database based on Gaussian Copulas \cite{Patki_2016}. One of the few publications treating multi-relation tables in their original form (to out knowledge the only), and has attracted a fair readership. See \href{https://github.com/sdv-dev/SDV}{Github sdv-dev/SDV}.},
    text={SDV},
    first={Synthetic Data Vault}
}

\newglossaryentry{pm}{
    type=\acronymtype,
    name={PM}, 
    description={According to the NIH, Precision Medicine or "Personalized medicine is an emerging practice of medicine that uses an individual's genetic profile to guide decisions made in regard to the prevention, diagnosis, and treatment of disease. Knowledge of a patient's genetic profile can help doctors select the proper medication or therapy and administer it using the proper dose or regimen." \cite{Ackerman2009}}
    text={PM},
    first={Personalized Medicine}
}

\newglossaryentry{st}{
    type=\acronymtype,
    name={ST}, 
    description={The self-training and co-training methods use classifiers first trained on the portion of labelled data to predict the labels of unlabelled instances. The newly labelled samples with the highest confidence are added to the labelled set to retrain the classifiers. The process is repeated iteratively. In the words of \citeauthor{yu2019rare}, \textit{"[...] a classifier is initially trained on the small labeled set, and the trained classifier is used to classify the unlabeled set, which is assigned with pseudo labels. After that, the part of unlabeled set with the most confident pseudo labels are selected, and added into the labeled set. The classifier iteratively trains itself with the labeled data and selected unlabeled data."} \cite{yu2019rare}.}
    text={ST},
    first={Self-training (ST)}
}

\newglossaryentry{ct}{
    type=\acronymtype,
    name={CT}, 
    description={The self-training and co-training methods use classifiers first trained on the portion of labelled data to predict the labels of unlabelled instances. The newly labelled samples with the highest confidence are added to the labelled set to retrain the classifiers. The process is repeated iteratively. In the words of \cite{yu2019rare}, \textit{"[...] co-training splits the features of labeled set into two sub-sets as two views, which are conditionally independent. Two classifiers are trained on two sub-sets respectively, and classify the unlabeled set with pseudo labeled. Then, the most confident unlabeled data determined by one classifier is fed into another classifier as additional pseudo labeled data for further training." \cite{yu2019rare}.}
    text={CT},
    first={Co-training (CT)}
}}

\newglossaryentry{mi}{
    type=\acronymtype,
    name={MI}, 
    description={See \gls{pd}.}
    text={Membership Inference},
    first={Membership Inference (MI)}
}

\newglossaryentry{pda}{
    type=\acronymtype,
    name={PD}, 
    description={Broadly, a Membership Inference attack aims to determine if a particular record was used to train a machine learning model \cite{chen2019ganleaks}. There is no canonical process by which an attack is conducted, nor specification of the data assets initially in possession of the attacker. Attacks range from completely \gls{fullbba} where the attacker can only query data from the model, to \gls{wba} where the model and its parameters all fully exposed. For a comprehensive taxonomy of MIA, refer to \citeauthor{chen2019ganleaks, Jayaraman2019}.}
    text={PD},
    first={Presence Disclosure (PD)}
}

\newglossaryentry{dle}{
    type=\acronymtype,
    name={DLE}, 
    description={Drug Laboratory Effects refer the changes that a patient's medication can induce on medical laboratory analyses such as diagnostic tests, leading to misinterpretations and errors \cite{VanBalveren2018}. Merely keeping track of the large quantity of known interactions is still problematic and the number of possible combination is immense. Moreover the effects vary according to each patient physiology. \citeauthor{yahi2017generative} made use of GANs to predict these effect on an personalized basis \cite{yahi2017generative}. See also \gls{ite}.} 
    text={DLE},
    first={Drug Laboratory Effects (DLE)}
}

\newglossaryentry{ite}{
    type=\acronymtype,
    name={ITE}, 
    description={Given a patient and what we know about the person's medical history and state, and the probability of various possible outcomes of disease progression. The aim of Individualized Treatment effects is to estimate the consequences of administering a particular treatment and their likely-hood. The task is made particularly complex due the fact that for any given individual, the decision can only be made once. In other words, paired samples are lacking, making it impossible to compare the outcomes directly \cite{Haupt2019}.}
    text={ITE},
    first={Individualized Treatment effects (ITE)}
}

\newglossaryentry{mimic}{
    type=\acronymtype,
    name={MIMIC}, 
    description={Openly available dataset of deidentified health data associated with ~60,000 intensive care unit admissions. It includes demographics, vital signs, laboratory tests, medications, a range of data types from physiological time-series to free-text interpretation of radiology imaging \cite{johnson2016mimic}}
    text={MIMIC},
    first={Medical Information Mart for Intensive Care (MIMIC)}
}

\newacronym[type=oalgo]{medgan}{medGAN}{medGAN}
\newacronym[type=oalgo]{ssl-gan}{SSL-GAN}{Semi-supervised Learning with a learned ehrGAN}
\newacronym[type=oalgo]{wgantpp}{PPWGAN}{\gls{wgan} for Temporal Point-processes}
\newacronym[type=oalgo]{radialgan}{RadialGAN}{RadialGAN}
\newacronym[type=oalgo]{mc-arae}{MC-ARAE}{Multi-categorical \gls{arae}}
\newacronym[type=oalgo]{ctgan}{CTGAN}{Conditional Tabular \Gls{gan}}
\newacronym[type=oalgo]{heterogan}{HGAN}{Heterogeneous GAN}
\newacronym[type=oalgo]{emr-wgan}{EMR-WGAN}{EMR Wassertein GAN}
\newacronym[type=oalgo]{corgan}{corGAN}{corGAN}
\newacronym[type=oalgo]{1d-cae}{1D-CAE}{1-dimensional Convolutional \gls{ae}}
\newacronym[type=oalgo]{ehrgan}{ehrGAN}{Electronic Health Record GAN}
\newacronym[type=oalgo]{rgan}{RGAN}{Recurrent \gls{gan}}
\newacronym[type=oalgo]{rcgan}{RC-GAN}{Recurrent Convolutional \gls{gan}}
\newacronym[type=oalgo]{ganite}{GANITE}{Generative Adversarial Nets for inference of Individualized Treatment Effects}
\newacronym[type=oalgo]{cwr-gan}{CWR-GAN}{Cycle Wasserstein Regression \gls{gan}}
\newacronym[type=oalgo]{gain}{GAIN}{Generative Adversarial Imputation Network}
\newacronym[type=oalgo]{cgain}{CGAIN}{Categorical \gls{gain}}
\newacronym[type=oalgo]{mc-medgan}{MC-medGAN}{Multi-categorical \gls{medgan}}
\newacronym[type=oalgo]{mc-gumbelgan}{MC-GumbelGAN}{Multi-categorical Gumbel-softmax \gls{gan}}
\newacronym[type=oalgo]{mc-wgan-gp}{MC-WGAN-GP}{Multi-categorical \gls{wgan} with Gradient Penalty}
\newacronym[type=oalgo]{medbgan}{MedBGAN}{Boundary-seeking \gls{medgan}}
\newacronym[type=oalgo]{healthgan}{HealthGAN}{HealthGAN}
\newacronym[type=oalgo]{medwgan}{MedWGAN}{Wassertein \gls{medgan}}
\newacronym[type=oalgo]{sc-gan}{SC-GAN}{Sequentially Coupled \gls{gan}}
\newacronym[type=oalgo]{rmb}{RMB}{Restricted Boltzmann Machine}
\newacronym[type=oalgo]{anomigan}{AnomiGAN}{GANs for anonymizing private medical data}
\newacronym[type=oalgo]{wgan-gp}{WGAN-GP}{\gls{wgan} with Gradient Penalty}
\newacronym[type=oalgo]{dp-auto-gan}{DP-auto-GAN}{\gls{dp}-auto-\gls{gan}}
\newacronym[type=oalgo]{ads-gan}{ADS-GAN}{Anonymization through data synthesis using \gls{gan}}
\newacronym[type=oalgo]{gcgan}{GcGAN}{\gls{corrnn} and \gls{t-wgan}}
\newacronym[type=oalgo]{t-wgan}{T-wGAN}{Wassertein \gls{t-gan}}
\newacronym[type=oalgo]{conan}{CONAN}{\textit{Co}plementary patter\textbf{n A}augmentatio\textbf{n}}
\newacronym[type=oalgo]{cwgan-gp}{cWGAN-GP}{Conditional \gls{wgan-gp}}
\newacronym[type=oalgo]{pate-gan}{PATE-GAN}{Private Aggregation of Teacher Ensembles (PATE) framework applied to GANs}
\newacronym[type=oalgo]{ac-gan}{AC-GAN}{Auxiliary Classifier \gls{gan}}
\newacronym[type=oalgo]{glugan}{GluGAN}{Blood Glucose \gls{gan}}
\newacronym[type=oalgo]{wgan-dp}{WGAN-DP}{\gls{wgan} with \gls{dp}}
\newacronym[type=oalgo]{rsdgm}{RSDGM}{Realistic Synthetic Dataset Generation Method}

\newacronym{nn}{NN}{Neural Network}
\newacronym{gan}{GAN}{Generative Adversarial Network}
\newacronym{ffn}{FFN}{Feed-forward Network}
\newacronym{ae}{AE}{Autoencoder}
\newacronym{rnn}{RNN}{Recurrent \gls{nn}}
\newacronym{lstm}{LSTM}{Long Short-term Memory}
\newacronym{cgan}{CGAN}{Conditional \gls{gan}}
\newacronym{crmb}{CRMB}{Conditional Restricted Boltzmann Machine}
\newacronym{cnn}{CNN}{Convolutional \gls{nn}}
\newacronym{wgan}{WGAN}{Wassertein \gls{gan}}
\newglossaryentry{beta-vae}{
    name = {0xCE-VAE},
    description = {0xCE variational auto-encoder}
}
\newacronym{lr}{LR}{Logistic-regression}
\newacronym{cycle-gan}{Cycle-GAN}{Cycle-consistent \gls{gan}}
\newacronym{adtep}{ADTEP}{Adversarial Deep Treatment Effect Prediction}
\newacronym{cae}{CAE}{Convolutional \gls{ae}}
\newacronym{gru}{GRU}{Gated Recurrent Unit}
\newacronym{gumbel-gan}{Gumbel-GAN}{Gumbel-Softmax \gls{gan}}
\newacronym{arae}{ARAE}{Adversarially regularized autoencoder}

\newacronym{ml}{ML}{Machine Learning}
\newacronym{ohd-gan}{OHD-GAN}{\glspl{gan} for Observation Health Data}
\newacronym{sd}{SD}{Synthetic Data}
\newacronym{ohd}{OHD}{Observational Health Data}
\newacronym{ehr}{EHR}{Electronic Health Record}
\newacronym{icu}{ICU}{Intensive Care Unit}
\newacronym{pmf}{PMF}{Probability Mass Function}
\newacronym{ssl}{SSL}{Semi-supervised learning}
\newacronym{cqm}{CQM}{Clinical Quality Measure}
\newacronym{hi}{HI}{Health Informatics}
\newacronym{pd}{PD}{Probability Distribution}
\newacronym{ks}{KS}{Kolmogorov-Smirnov}

\newacronym{iot}{IoT}{Internet of Things}

\newacronym{bn}{BN}{batch-normalization}
\newacronym{sc}{SC}{shortcut connections}
\newacronym{cbt}{CBT}{Cluster-based training}
\newacronym{vcd}{VCD}{Variational contrastive divergence}
\newacronym{ln}{LN}{Layer normalisation}
\newacronym{sn}{SN}{Spectral Normalization}
\newacronym{dsm}{DSM}{Domain Specific Measure}
\newacronym{dwp}{DWP}{Dimension-wise prediction}
\newacronym{arm}{ARM}{Association Rule Mining}
\newacronym{trts}{TRTS}{Train on synthetic, test on real}
\newacronym{tstr}{TSTR}{Train on real, test on synthetic}
\newacronym{dp}{DP}{Differential privacy}
\newacronym{dp-sgd}{DP-SGD}{Differential private stochastic gradient descent}
\newacronym{ad}{AD}{Attribute Disclosure}
\newacronym{rr}{RR}{Reproduction rate}

\newacronym{anm}{ANM}{Additive noise model}
\newacronym{vae}{VAE}{Variational AE}
\newacronym{fop}{F-OP}{First-order proximity}
\newacronym{cc}{CC}{Correlation coefficient}
\newacronym{mmd}{MMD}{Maximum Mean Discrepency}
\newacronym{rbf}{RBF}{Radial Basis Function}
\newacronym{mse}{MSE}{Mean Squared Error}
\newacronym{auroc}{AUROC}{Area under ROC curve}
\newacronym{auprc}{AUPRC}{Area under the precision-recall curve}
\newacronym{kld}{KLD}{Kullback-Leibler divergence}
\newacronym{fd}{FD}{Feature distributions}
\newacronym{qq}{QQ}{Quantile-quantile plot}
\newacronym{lsr}{LSR}{Latent space representation}
\newacronym{rdp}{RDP}{Renyi Differential Privacy}
\newacronym{pcam}{PCAM}{\gls{pca} Marginal}
\newacronym{pca}{PCA}{Principal Component Analysis}
\newacronym{pcawdd}{PCA-DWD}{\gls{pca} Distributoinal Wassertein Distance}
\newacronym{pta}{PTA}{Prediction task accuracy}
\newacronym{ssa}{SSA}{Semi-supervised augmentation}
\newacronym{dwpro}{DWS}{Dimension-wise Statistics}
\newacronym{dwpre}{DWP}{Dimension-wise Prediction}
\newacronym{ved}{VED}{Visual Expert Discrimination}
\newacronym{rf}{RF}{Random Forest}
\newacronym{svm}{SVM}{Support Vector Machine}
\newacronym{rmse}{RMSE}{Root Mean-Squared Rrror}
\newacronym{rvts}{RV-TS}{real-valued time-series}
\newacronym{dts}{D-TS}{discrete time-series}
\newacronym{vs}{VS}{Variable Splitting}
\newacronym{it}{IT}{Iterative Imputation}
\newacronym{bp}{BP}{Backpropagation \gls{it}}
\newacronym{pr}{PR}{Posterior Regularization}
\newacronym{rl}{RL}{Reinforcement learning}
\newacronym{ps}{PS}{Posterior Regularization}
\newacronym{irl}{IRL}{Inverse \gls{rl}}
\newacronym{hexagan}{HexaGAN}{Six component \gls{gan}}
\newacronym{miwae}{MIWAE}{Missing data IWAE}
\newacronym{hi-vae}{HI-VAE}{Heterogeneous-Incomplete VAE}
\newacronym{mida}{MIDA}{Multiple Imputation Denoising Autoencoders}
\newacronym{info-gan}{InfoGAN}{Information Maximizing \gls{gan}}
\newacronym{eeg}{EEG}{Electroencephalogram}
\newacronym{ecg}{ECG}{Electrocardiogram}

\title{ Synthetic Observational Health Data with GANs: from slow adoption to a boom in medical research and ultimately digital twins?}
\author{
  Georges-Filteau, Jeremy \href{https://orcid.org/0000-0002-0352-6468}{\includegraphics[width=10pt]{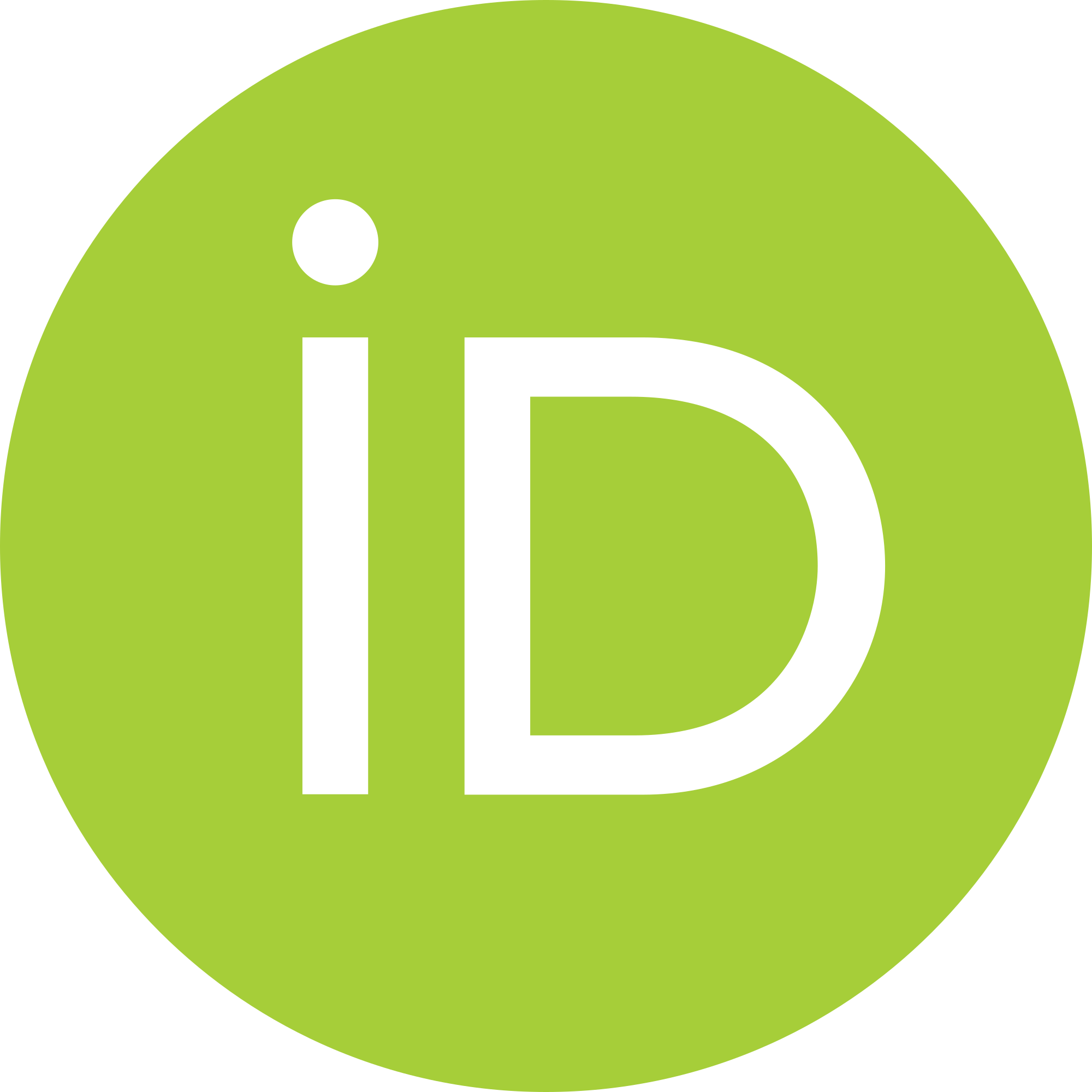}}\\[0.2cm]
  \small \textit{Radboud University, The Hyve}\\
  \small\texttt{jeremy@thehyve.nl}
  \and
  Cirillo, Elisa \href{https://orcid.org/0000-0002-0241-7833}{\includegraphics[width=10pt]{assets/orcid.png}}\\[0.2cm]
  \small \textit{The Hyve}\\
  \small\texttt{elisa@thehyve.nl}
}

\begin{document}

    \maketitle
    \vspace{-1em}

    \begingroup
    \let\center\flushleft
    \let\endcenter\endflushleft
    \maketitle
    \endgroup

    \selectlanguage{english}

    \glsresetall
    \begin{abstract}
    After being collected for patient care, \gls{ohd} can further benefit patient well-being by sustaining the development of health informatics and medical research. Vast potential is unexploited because of the fiercely private nature of patient-related data and regulation about its distribution. 
    \glspl{gan} have recently emerged as a groundbreaking approach to learn generative models efficiently that produce realistic \gls{sd}. They have revolutionized practices in multiple domains such as self-driving cars, fraud detection, simulations in the and marketing industrial sectors known as digital twins, and medical imaging. The digital twin concept could readily apply to modelling and quantifying disease progression. In addition, \glspl{gan} posses a multitude of capabilities relevant to common problems in the healthcare: augmenting small dataset, correcting class imbalance, domain translation for rare diseases, let alone preserving privacy. Unlocking open access to privacy-preserving \gls{ohd} could be transformative for scientific research. In the COVID-19's midst, the healthcare system is facing unprecedented challenges, many of which of are data related and could be alleviated by the capabilities of \glspl{gan}.
   Considering these facts, publications concerning the development of  \gls{gan} applied to \gls{ohd} seemed to be severely lacking. To uncover the reasons for the slow adoption of \glspl{gan} for \gls{ohd}, we broadly reviewed the published literature on the subject. Our findings show that the properties of \gls{ohd} and evaluating the \gls{sd} were initially challenging for the existing \gls{gan} algorithms (unlike medical imaging, for which state-of-the-art model were directly transferable) and the choice of metrics ambiguous. We find many publications on the subject, starting slowly in 2017 and since then being published at an increasing rate. The difficulties of \gls{ohd} remain, and we discuss issues relating to evaluation, consistency, benchmarking, data modeling, and reproducibility.
    \end{abstract} 

    \glsresetall
\section{Introduction}
    \subsection{Background}
        Medical professionals collect \gls{ohd} in \glspl{ehr} at various points of care in a patient’s trajectory, to support and enable their work \cite{Cowie_2016}. The patient profiles found in \glspl{ehr} are diverse and longitudinal, composed of demographic variables, recordings of diagnoses, conditions, procedures, prescriptions, measurements and lab test results, administrative information, and increasingly omics \cite{Ohdsi2020-vf}.\par
        Having served its primary purpose, this wealth of detailed information can further benefit patient well-being by sustaining medical research and development. That is to say, improving the development life-cycle of \gls{hi}, the predictive accuracy of \gls{ml} algorithms, or enabling discoveries in research on clinical decisions, triage decisions, inter-institution collaboration, and \gls{hi} automation \cite{Rudin_2020, Rankin2020}. Big health data is the underpinning of two prime objectives of precision medicine: individualization of patient interventions and inferring the workings of biological systems from high-level analysis \cite{Capobianco2020}. However, the private nature of patient-related data and the growing widespread concern over its disclosure hampers dramatically the potential for secondary usage of \gls{ohd} for legitimate purposes.\par
        Anonymization techniques are used to hinder the misuse of sensitive data. This implies a costly and data-specific cleansing process, and the unavoidable trade-off of enhancing privacy to the detriment of data utility \cite{Dankar2012-bd, Cheu2019-vh, De_Cristofaro2020-tl}. These techniques are fallible and do not prevent reidentification. In fact, no polynomial time \gls{dp} algorithms can produce \gls{sd} preserving all relations of the real data, even for simple relations such as 2-way marginals \cite{Ullman2011}. To address these drawbacks, alternative modes for sharing sensitive data is an active research area, including privacy-preserving analytic and distributed learning. Although promising, these approaches come with limitations, and we must still explore their feasibility and scalability \cite{Raisaro2018-gv}. Regardless, distributed models are vulnerable to a variety of attacks, for which no single protection measure is sufficient as research on defense is far behind attack \cite{enthoven2020overview, Gao2020, Luo2020-gq, Lyu2020-sv}.\par
        These conditions restrict access to \gls{ohd} to professionals with academic credentials and financial resources. Use of OHD by all other health data-related occupations is blocked, along with the downstream benefits. For example, software developers rarely have access to the data at the core of the \gls{hi} solutions they are developing, or educators lack examples \cite{laderas_teaching_2018}.
        
    \subsection{Synthetic data}
        An alternative to traditional privacy-preserving methods is to produce full \gls{sd}. We categorize methods to produce \gls{sd} as either theory-driven (theoretical, mechanistic or iconic) or data-driven (empirical or interpolatory) modelling \cite{Kim_2017, Hand2019}. Theory-driven modelling involves a complex knowledge-based attempt to define a simulation process or a statistical model representing the causal relationships of a system \cite{Yousefi2018-dy, Kansal2018-dx}. The Synthea \cite{Walonoski_2017} synthetic patient generator is one such model, in which state transition models\footnote{Probabilistic model composed of pre-defined states, transitions, and conditional logic.} produce patient trajectories. It derives the model parameters from aggregate population-level statistics of disease progression and medical knowledge. Such a knowledge-based model depends on prior knowledge of the system, and how much we can intellect about it \cite{Kim_2017, Bonnery2019-ug}. On one hand, theory-based modelling aims at understanding and offers interpretability, on the other when modelling complex systems, simplifications and assumptions are inevitable, leading to inaccuracies or reduced utility \cite{Hand2019, Rankin2020}. In fact, relying on population-level statistics does not produce models capable of reproducing heterogeneous health outcomes \cite{Chen_2019}.\par
        
        Data-driven modelling techniques infer a representation of the data from a sample distribution, to summarize or describe it \cite{Hand2019}. There are many statistical modelling approaches to produce \gls{sd}, but intrinsic assumptions about the data form the basis. They bound their representational power to correlations intelligible to the modeler, being prone to obscure inaccuracies. \gls{sd} generated by these models hits a ceiling of utility \cite{Rankin2020}. In the ML field, generative models learn an approximation of the multi-modal distribution, from which we can draw synthetic samples  \cite{goodgan}. \Gls{gan} \cite{goodgan} have recently emerged as a groundbreaking approach to learn generative models that produce realistic \gls{sd} using \gls{nn}. \gls{gan} algorithms have rapidly found a wide range of applications, such as data augmentation in medical imaging \cite{Yi2019, Wang2020, Zhou2020}.\par
        
        The potential affects of \gls{gan} to healthcare and science are considerable \cite{Rankin2020}, some of which have been realized in fields such as medical imaging. However, the application of \gls{gan} to \gls{ohd} seems to have been lagging \cite{Xiao_2018_chall}. Well-known characteristics of \gls{ohd} could explain the relatively slow progress. Primarily, algorithms developed for images and text in other fields were easily re-purposed for medical equivalents of the data types. However, \gls{ohd} presents a unique complexity in terms of multi-modality, heterogeneity, and fragmentation \cite{Xiao_2018_chall}. In addition, evaluating the realism of synthetic \gls{ohd} is intuitively complex, a problem that still burdens \glspl{gan}. In 2017, a few authors first attempts at \glspl{gan} for \gls{ohd} were published \cite{esteban2017real,Che_2017,Choi2017-nt,yahi2017generative}. We aimed to investigate if these examples inspired more research, and if so, to gain a comprehensive understanding of approaches to the problem and the techniques involved.

    \section{Methods}

    \begin{table}[h]
  \center
  \footnotesize
    \caption{Search query terms}\label{tab:search}
    \begin{tabular}{@{}clccl@{}} \toprule
	    \multicolumn{2}{c}{Health data} & & \multicolumn{2}{c}{Generative adversarial models} \\ \cmidrule{1-2} \cmidrule{4-5}
	    \multicolumn{2}{c}{Terms} & {} & \multicolumn{2}{c}{Terms} \\ \cmidrule{2-2} \cmidrule{5-5}
	    \multirow{4}{*}{OR} & clinical & \multirow[t]{4}{*}{\quad AND\quad} & \multirow{4}{*}{OR} & generative adversarial\\
	    {} & health & {} & {} & GAN \\ 
	    {} & EHR & {} & {} & adversarial training \\
	    {} & electronic health record & {} & {} & synthetic  \\
	    {} & patient & {} & {} & {} \\
	    \bottomrule
    \end{tabular}
\end{table}
    
    Publications concerning \gls{ohd-gan} were identified through with Google Scholar \cite{scholar}, Web of Science \cite{Clarivate} and Prophy \cite{Prophy}. The terms and operators found in Table \ref{tab:themes} form the search query. We included studies reporting the development, application, performance evaluation and privacy evaluation of \gls{gan} algorithms to produce \gls{ohd}. We define \gls{ohd} as categorical, real-valued, ordinal or binary event data recorded for patient care. We list a more detailed summary of the included and excluded data types in Table \ref{tab:datatypes}. The excluded data types are already the subject of one or more review, or would merit a review of their own \cite{Yi_2019, Nakata2019, Anwar_2018, Wang2020, Zhou2020}. In each of the included publications, we considered the aspects listed in Table \ref{tab:search}.\par

        \begin{table}[h]
            \centering
            \footnotesize
              \caption{Aspects analysed in each of the publications included in the review\label{tab:themes}}
              \begin{tabular}{ll}\toprule
              A) Types of healthcare data & D) Evaluation metrics\\
              B) \gls{gan} algorithm, learning procedures, losses & E) Privacy considerations\\
              C) Intended use of the \gls{sd} & F) Interpreatability of the model\\\bottomrule
              \end{tabular}
        \end{table}

    \begin{table}[htb]
\center
\footnotesize
  \caption{Types of OHD data included or excluded from the review.}\label{tab:datatypes}
  
  \begin{tabularx}{\textwidth}{@{} p{0.1\textwidth}Xp{0.7\textwidth}@{}}\toprule
  Type & Examples \\ \midrule
  
  \multirow{4}{*}{Included} & Observations & Demographic information, medical classification, family history \\
  
  &Time-stamped observations & Diagnosis, treatment and procedure codes, prescription and dosage, laboratory test results, physiologic measurements and intake events \\
  &Encounters & Visit dates, care provider, care site \\
  &Derived & Aggregated counts, calculated indicators. \\ \midrule

  \multirow{4}{*}{Excluded} & Omics & Genome, transcriptome, proteome, immunome, metabolome, microbiome \\
  &Imaging & X-rays, computed tomography (CT), magnetic resonance imaging (MRI) \\
  &Signal & Electrocardiogram (ECG), electroencephalogram (EEG) \\
  &Unstructured & Narrative reports, textual \\ \bottomrule
  \end{tabularx}%
\end{table}

    \section{Results}
    \subsection{Summary}
        We found 43 publications describing the development or adaption of \gls{ohd-gan}, presented in Table \ref{tab:3:publications}. We can generalize the data addressed in each of these publications into one of two categories: time-dependent observations, such as time-series, or static representation in the form of feature vectors such as tabular rows. We briefly bring attention to the lack of multi-relational tabular representations, the primary form of \gls{ehr}, and further discuss the subject in latter sections.\par
        
        Most efforts propose adaptations of current algorithms to the characteristics and complexities of \gls{ohd}. These include multi-modality of marginal distributions or non-Gaussian real-valued features, heterogeneity, a combination of discrete and real-valued features, longitudinal irregularity, complex conditional distributions, missingness or sparsity, class imbalance of categorical features and noise.\par 
        
        While these properties may make training a useful model difficult, the variety of applications that are highly relevant and needed in the healthcare domain provides sufficient incentive. The most cited motives are, as one would expect, to cope with the often limited number of samples in medical datasets and to overcome the highly restricted access to \gls{ohd}. The potential of releasing privacy-preserving \gls{sd} freely is a common subject. Publications considering privacy evaluate the effect on utility of applying \gls{dp} to their algorithm, propose alternatives privacy concepts and metrics, or only concentrate on the subject of privacy.\par
        
    \subsection{Motives for developing OHD-GAN}
        Some claim that the ability to generate synthetic is becoming an essential skill in data science \cite{Sarkar2018}, but what purpose can it serve in the medical domain? The authors mention a wide range of potential applications. We briefly describe the four prevailing themes in the following sections: data augmentation (Sec.\ref{sec:augmentation}), privacy and accessibility (Sec.\ref{sec:access_privacy}), precision medicine (Sec.\ref{sec:precision_med}) and  modelling simulations (Sec.\ref{sec:models_twins}). 

        \subsubsection{Data augmentation}\label{sec:augmentation}
    
            Data augmentation \todo{define} is mentioned in nearly all publications. Although counter-intuitive, \gls{gan} can generate \gls{sd} that conveys more information about the real data distribution. Effectively, the real-valued space distribution of the generator produces a more comprehensive set of data points, valid, but not present in the discrete real data points. A combination of real and synthetic training data habitually leads to increased predictor performance \cite{Wang_2019,Che_2017,Yoon2018-ite, yoon2018imputation, Yang_2019_impute_ehr, Chen_2019, cui2019conan, Che_2017}. A more intelligible way would be to seize the concept from the point of view of image classification, known as invariances, perturbations such as rotation, shift, sheer and scale \cite{antoniou2017data}.\par 
            
            Similarly, domain translation \todo{define} and \gls{semi-sup} training approaches with \glspl{gan} could support predictive tasks that lack data with accurate labels, lack paired samples or suffer class imbalance \cite{Che_2017,mcdermott2018semi, Yoon2018-ite}. Another example is correcting discrepancies between datasets collected in different locations or under different conditions inducing bias \cite{Yoon2018-radial}. \glspl{gan} are also well adapted for data imputation, were  entries are \gls{mar} \cite{yoon2018imputation}. 

        \subsubsection{Enhancing privacy and increasing data accessibility}\label{sec:access_privacy}
    
            Most authors see \gls{sd} as the key to unlocking the unexploited value of \gls{ohd} hindering machine learning, and scientific progress \cite{Beaulieu-Jones2019-ct, baowaly_2019_IEEE,baowaly_2019_jamia,Che_2017,esteban2017real,Fisher2019,severo2019ward2icu} or education \cite{laderas_teaching_2018}. We can broadly describe preserving privacy as reducing the risk of \gls{re-iden} to an acceptable level. It quantifies this level of risk when releasing data anonymized with \gls{dp}.\par
    
            Due to its artificial nature, \gls{sd} is put forward to forgo the tight restrictions on data sharing, while potentially providing greater privacy guarantees \cite{Beaulieu-Jones2019-ct, baowaly_2019_IEEE, baowaly_2019_jamia,esteban2017real,Fisher2019,walsh2020generating, chin2019generation}. Enabling access to greater variety, quality and quantity of \gls{ohd} could have positive effects in a wide range of fields, such as software development, education, and training of medical professionals. The fact remains that \glspl{gan} do not eliminate the risk of reidentification. Considering none of the synthetic data points represent actual people, the significance of such an occurrence is unclear. It is possible to combine both methods, and \gls{gan} training according to \gls{dp} shows evidence of reducing the loss of utility compared to \gls{dp} alone.
    
        \subsubsection{Enabling precision medicine}\label{sec:precision_med}
    
            The application to precision medicine involves predicting outcomes conditioned on a patient's current state and history. Simulated trajectories could help inform clinical decision making by quantifying disease progression and outcomes and have a transformative effect on healthcare \cite{walsh2020generating, Fisher2019}. Ensembles of stochastic simulations of individual patient profiles such as those produced by \gls{crmb} could help quantify risk at an unprecedented level of granularity \cite{Fisher2019}.\par
            Predicting patient-specific responses to drugs is still a new field of research, a problem known as \gls{ite}. Estimating \glspl{ite} is persistently hampered by the lack of paired counterfactual samples \cite{Yoon2018-ite, chu2019treatment}. To solve similar problems in medical imaging, various \gls{gan} algorithms were developed for domain translation, mapping a sample from its to original class to the paired equivalent. This includes bidirectional transformations, allowing \gls{gan} to learn mappings from very few, or a lack of paired samples \cite{Wolterink2017DeepMT, CycleGAN2017, mcdermott2018semi}.
    
        \subsubsection{From patient and disease models to digital twins}\label{sec:models_twins}
    
            A well-trained model approximates the process that generated the real data points. The relations learned by the model, its parameters, contain meaningful information if we can learn to harness it. Data-driven algorithms evolve as our understanding of their behavior improves. We incorporate new concepts in the algorithms leading to further understanding, interactivly blurring the line with theory-driven approaches \cite{Hand2019}. Interpretability is a growing field of research concerned with understanding how the learned parameters of a model relate. In other words analysing the representation the algorithm has converged to and deriving meaning from obscure logic. Incorporating new understanding in the architecture of algorithms shifts the view from a data-driven to a theory-driven perspective \cite{Hand2019}. As we purposefully build structure in our algorithms from new understanding, we may get the chance to explore meaningful representations that would otherwise be beyond our reasoning.\par 
            
            Approaching these ideas from above, the concept of "digital twins" represents in a way the ultimate realization of \gls{pm}. A common practice in industrial sectors is high-fidelity virtual representations of physical assets. Long-term simulations, that provide an overview and comprehensive understanding of the workings, behavior and life-cycle of their real counterparts. The state of the models is continuously updated from theoretical data, real data, and streaming \gls{iot} indicators.\par
            Intently conditioned input data allows the exploration of specific events or conditions. In a position paper on the subject, Angulo et al. draw the parallels of this technique with the current needs in healthcare and the emergence of the technologies for actionable models of patients. \cite{angulo2019towards,Angulo_2020}. The authors bring up the rapid adoption of wearables that are continuously monitoring people's physiological state.\par 
            
            Wearables are one of many mobile digitally connected devices that collect patient data over a broad range of physiological characteristic and behavioral patterns \cite{coravos2019developing}. This emerging trend known as \gls{dbio} has already led to studies demonstrating predictive models with the potential for improved patient care \cite{snyder2018best}. Through continuous lifelong learning, integrating  multiple modes of personal data, generative patient models could inform diagnostics of medical professionals and also enable testing treatment options. In their proposal, \gls{gan} are an essential component of the ecosystem to ensure patient privacy and to provide bootstrap data. Fisher et al. already use the term "digital twin" to describe their process, noting that they present no privacy risk and enable simulating patient cohorts of any size and characteristics \cite{walsh2020generating}.

\newcolumntype{R}{>{\raggedright\arraybackslash}p{0.20\textwidth}}
\newcolumntype{M}{>{\raggedright\arraybackslash}p{0.40\textwidth}}
\newcolumntype{N}{>{\raggedright\arraybackslash}p{0.30\textwidth}}
\newcommand{\specialcell}[2][c]{%
  \begin{tabular}[#1]{@{}l@{}}#2\end{tabular}}
  
\begin{center}
    
    \setlength\LTleft{0pt}
    \setlength\LTright{0pt}
    \scriptsize
    \setlength{\extrarowheight}{1em}
    
    \begin{longtable}[l]{@{}p{0.10\textwidth}NMR@{}} 
        \kill
        \caption{Summary of the publication included in the review.\label{tab:3:publications}}\\
        \hline
        Publication & Algorithm(s) & Focus, algorithms, and techniques & Data type \\ 
        \hline
        \endfirsthead
        \caption[]{Summary of the publication included in the review (Continued).}\\
        \hline
        Publication & Algorithm(s) & Focus, algorithms and techniques & Data \\ 
        \hline
        \endhead
        \hline 
        \endfoot
        
        \quad 2017 & & & \\
        \hline
        \citeauthor{Choi2017-nt} & \gls{medgan} 
        & Incompatibility of back-propagation with discrete features. \gls{ae}, \gls{mb-avg}, \gls{bn}, \gls{sc}, \gls{ad}, \gls{pda}.
        & Binary occurences or counts of medical codes.\\
        \citeauthor{yahi2017generative} & \gls{medgan} adaptation
        & \Gls{dle} on continuous time-series, multi-modality. \gls{t-sne}.
        & Paired pre/post treatment exposure time-series\\
        \citeauthor{esteban2017real} & \gls{rgan}, \gls{rcgan} 
        &  Adversarial training of (conditional) \glspl{rnn} on time-series, evaluation, privacy. \gls{lstm}, \gls{cgan}, \gls{dp-sgd}.
        & Regularly observed \gls{rvts}\\
        \citeauthor{Xiao2017-lh} & \gls{wgantpp} 
        & Temporal Point Processes. \gls{lstm}, \gls{wgan}, Poisson process.
        & Sporadic occurrences, hospital visits.\\
        \citeauthor{Che_2017} & \gls{ehrgan}, \gls{ssl-gan} 
        & Semi-supervised augmentation, transitional distribution. 1D-CNN, Word2vec, \gls{vcd}.
        & \Gls{dts}, sequences of medical codes. \\
        \citeauthor{dash2019synthetic} & \gls{healthgan} & Sleep patterns, stratification by covariates. & Binary over multiple visits.\\
        
        \hline
        \quad 2018 & & & \\
        \hline
        \citeauthor{Camino2018-re} & \raggedright \gls{mc-arae}, \gls{mc-medgan}, \gls{mc-gumbelgan}, \gls{mc-wgan-gp}
        & Improving training process. \gls{medgan}, \gls{wgan-gp}, \gls{gumbel-gan}, \gls{arae}.
        & Multiple categorical variables. \\
        \citeauthor{mcdermott2018semi} & \gls{cwr-gan}
        & Cycle-consistent semi-supervised regression learning, unpaired data, class imbalance. \gls{wgan} \gls{cycle-gan} \gls{ite}
        & ICU \gls{rvts}, lack of paired samples, \gls{sd}. \\
        \citeauthor{Yoon2018-ite} & \gls{ganite} 
        & \gls{ite}, unobserved counterfactual, multi-label classification, uncertainty. \gls{cgan} pair.
        & Feature, treatment and outcome vectors.\\
        \citeauthor{Yoon2018-radial} & \gls{radialgan} 
        & Multi-domain translation, features and distribution mismatch, cycle-consistency, augmentation. \gls{cgan}, \gls{wgan}.
        & Tabular, discrete and continuous.\\
        \citeauthor{yoon2018imputation} & \gls{gain}
        & Tabular data imputation. \gls{mcar}, \gls{cgan}.
        & Real-valued, tabular with entries \gls{mcar}.\\
        
        \hline
        \quad 2019 & & & \\
        \hline
        \citeauthor{Wang_2019} & \gls{sc-gan}
        & Capturing mutual influence in time-series. Coupled generator pair. Treatment recommendation task. \gls{lstm}, \gls{cgan}.
        & \Gls{rvts} of patient state and medication dosage data.\\
        \citeauthor{baowaly_2019_IEEE} & \gls{medbgan}
        & Improving training process. \gls{medgan}, \gls{bgan}.
        & Binary occurences or counts of medical codes.\\
        \citeauthor{baowaly_2019_jamia} & \gls{medbgan}, \gls{medwgan}
        & Improving training process. \gls{medgan}, \gls{bgan}, \gls{wgan}.
        & Binary occurences or counts of medical codes.\\
        \citeauthor{severo2019ward2icu} & \gls{cwgan-gp} 
        & Generation and public release of dataset. Protecting commercial sensitive information. Class imbalance. \gls{cwgan-gp}, \gls{cgan}.
        & Physiological \gls{rvts}.\\
        \citeauthor{chin2019generation} & \gls{wgan}
        & Heterogeneous mixture of dense and sparse features. Privacy and evaluating the introduction of bias. \gls{wgan}, \gls{wgan-gp}, \gls{msn}, \gls{dp} aware optimizer from Tensor-flow. \citeauthor{tensorflow-privacy}.
        & Binary, real-valued and categorical.\\
        \citeauthor{Jordon2019} & \gls{pate-gan}
        & Alternative differential privacy, adaptation of the  \gls{pate} framework.
        & Demographic and binary.\\
        \citeauthor{torfi2019generating} & \gls{corgan}
        & \gls{cnn} architecture, capturing feature correlations, evaluating realism, privacy evaluation using \gls{mi}. 1D-\gls{cae}.
        & Binary occurences or counts of medical codes.\\
        \citeauthor{chu2019treatment} & \gls{adtep}
        & \gls{ite}, two independent \gls{ae} for patient and treatment feature sets, trained adversarially in combination, and outcome predictor from latent representation. &
         \Gls{ehr} data, not specified.\\
        \citeauthor{Jackson_2019} & \gls{medgan}
        & Evaluating medgan with the addition of demographics features.
        & Demographic features and binary occurences or counts of medical codes. \\
        \citeauthor{yu2019rare} & \gls{ssl-gan}
        & Rare disease detection, \gls{ssl}, leveraging unlabeled \gls{ehr} data, medical code embedding network.  \gls{lstm}.
        & Diagnosis and prescription codes.\\
        \citeauthor{Yang_2019_cdss} & \gls{cgan}
        & Class imbalance, low count of minority class. Semi-supervised learning combining \gls{st} and \gls{ct} with a \gls{cgan} for a \gls{iot} application.
        & Twenty medical datasets from the UCI repository, types unspecified.\\
        \citeauthor{Yang_2019_ehr} & \gls{gcgan}
        & Capturing the correlations between different categories of medical codes and the outcome.  \gls{corrnn}, \gls{t-gan}, \gls{t-wgan}.
        & Binary occurences or counts of medical codes.\\
        \citeauthor{Yang_2019_impute_ehr} & \gls{cgain}
        & Improve on \gls{gain} for categorical variable using fuzzy encoding of the features. 
        & Categorical (multi-class and multi-label) real-valued.\\
        \citeauthor{Camino2019} & \gls{gain}, \gls{gain}+\gls{vs}, \gls{vae}, \gls{vae}+\gls{it}, \gls{vae}+\gls{bp},  \gls{vae}+\gls{vs},  \gls{vae}+\gls{vs}+\gls{it}, \gls{vae}+\gls{vs}+\gls{bp}
        & Benchmark and improve on generative imputation with \gls{gain} and \gls{vae}. & Categorical and real-valued. Mostly not \gls{ohd}\\
        \citeauthor{Beaulieu-Jones2019-ct} & \gls{ac-gan} 
        & Evaluating if differentially private GANs that is valid reanalysis while ensuring privacy.  \gls{dp}, \gls{cgan}.
        & Physiological \gls{rvts}.\\
        \citeauthor{Xu2019-ay} & \gls{ctgan}
        & Non-Gaussian multi-modal distribution of continuous columns and imbalanced discrete column in tabular data. Evaluation benchmark.  \gls{cgan} \gls{tbs} \gls{msn} \gls{wgan-gp} \gls{gumbel-gan}
        & Tabular real-valued and categorical.\\
        \citeauthor{yale2019ESANN} & \gls{healthgan}
        & Privacy metrics and over-fitting.  \gls{mi}, \gls{nnaa}, \gls{pl}, \gls{dt}
        & Categorical demographics, real-valued and binary medical codes.\\
        \citeauthor{Fisher2019} & Adversarially trained \gls{crmb}
        & Simulation of patient trajectories from their baseline state, disease prediction and risk quantification, missingness.\gls{crmb}.
        &  Binary, ordinal, categorical, and continuous, 3 months intervals.\\
        
        \hline
        \quad 2020 & & & \\
        \hline
        \citeauthor{walsh2020generating} & Adversarially trained \gls{crmb}
        & Digital twins, disease prediction and risk quantification, missingness.   \gls{crmb}.
        & Binary, ordinal, categorical, and continuous, 3 months intervals.\\
        \citeauthor{Yale_2020} & \gls{healthgan}
        & Metrics to capture a synthetic dataset’s resemblance, privacy, utility and footprint. Evaluating applications. Application case studies, Reproducibility of studies with \gls{sd}.  \gls{nnaa}, \gls{pl}, \gls{do}, \gls{medgan}, \gls{wgan-gp}, \gls{sdv}, 
        & Real-valued and categorical. Demographics, vital signs, diagnoses, and procedures.\\
        \citeauthor{tanti2019} & \gls{dp-auto-gan}
        & Privacy, \gls{medgan} adaptation, evaluation metrics.  \gls{dp-sgd} \gls{ae} \gls{medgan} \gls{rdp}
        & Medical data: binary. Non-health data: categorical and real-valued.\\
        \citeauthor{BaeAnomiGAN2020} & \gls{anomigan}
        & Probabilistic scheme that ensures \textit{indistinguishability} of the \gls{sd} can be viewed as encrypted.  \gls{dp} \gls{cnn}
        & Binary occurences of medical codes.\\
        \citeauthor{cui2019conan} & \gls{conan}
        & Complementary \gls{gan} in a rare disease predictor model that generates positive samples from negatives to alleviate class imbalance.
        & Embedding vectors representing multiple patient visits and conditions.\\
        \citeauthor{zhu_2020} & \gls{glugan}
        & Adversarialy trained \gls{rnn} to predict the upcoming time-step in physiological time-series conditioned on the past observations.  \gls{rnn}, \gls{cnn}, \gls{gru}.
        & \Gls{rvts} of blood glucose measurements, discrete patient submitted features.\\
        \citeauthor{chen2019ganleaks} & \gls{medgan}, \gls{wgan-gp}, DC-GAN
        & Privacy analysis of generative models.   \gls{mi}, \gls{fullbba}, \gls{partbba}, \gls{wba}, \gls{dp-sgd}.
        & Binary vector of medical codes.\\
        \citeauthor{chincheong2020generation} & \gls{wgan-dp}
        & Heterogeneous data, effect of differential privacy on utility.   \gls{wgan} \gls{dp}
        & Categorical, continuous,  ordinal, and binary. Dense or sparse.\\
        \citeauthor{Camino2020bench} & - & Initially a comparison \glspl{gan} and \glspl{vae}, but they choose instead to bring attention to the problem of benchmarking. Analysis of problematic,  requirements and suggestions. \gls{gain}, \gls{hexagan} \cite{hwang2019exagan}, \gls{miwae} \cite{mattei2019miwae}, \gls{hi-vae}\cite{nazabal2020handling}, \gls{mida} \cite{Gondara2017} & Real-valued and categorical.\\
        \citeauthor{Zhang2020} & \gls{emr-wgan}
        & Improving training, evaluation metrics, sparsity.  \gls{wgan},\gls{bn},\gls{ln}, \gls{cgan}.
        & Binary occurences of medical codes. Low-prevalence of codes. \\
        \citeauthor{yan2020generating} & \gls{heterogan}
        & Improvements on \gls{emr-wgan} incorporating record-level constraints in the loss function.   \gls{wgan}, \gls{bn}, \gls{ln}, \gls{cgan}, \gls{mi}, \gls{pda}.
        & Binary, categorical and real-valued.\\
        \citeauthor{ozyigit2020generation} & \gls{rsdgm}
        & Exploring the feasibility of various methods to generate synthetic datasets.   \gls{wgan}
        & Real-valued and categorical.\\
        \citeauthor{Yoon2020-anon} & \gls{ads-gan}
        & Identifiability view of privacy. Generator conditioned on real samples inputs with an identifiability loss to satisfy the identifiability constraint.   \gls{wgan} \gls{wgan-gp} \gls{dp} alternative.
        & Real-valued and binary.\\
        \citeauthor{Goncalves2020} & \gls{mc-medgan}
        & Comparison of \glspl{gan} with statistical models to generate synthetic data, evaluation metrics.   \gls{mi}, \gls{ad}.
        & Categorical and real-valued.\\
           
        \hline
        
    \end{longtable}
\end{center}

    \subsection{Data Types and Feature Engineering}

        No publications made use of \gls{ohd} in its initial form, patient records in \gls{ehr} are composed of many related tables (normalized form). The complexity of a model would explode when maintaining referential integrity and statistics between multiple tables. The hierarchy by which these would interact with each other conditionally is no less complicated \cite{Buda2015, Patki_2016, Zhang2015, Tay2013}. There are published \gls{gan} algorithms made to consume normalized database in their original form. In all publications we considered, feature engineering was used to adapt the data to task requirements, or to promising algorithms that fit the data characteristics. They transform the data into one of four modalities: time series, point-processes, ordered sequences or aggregates described in Fig. \ref{tab:features}.

        \begin{table}[H]
        \footnotesize
        \caption{Types of observational health data and features engineering}\label{tab:features}
    
        \begin{tabularx}{\textwidth}{@{}p{0.15\textwidth}p{0.3\textwidth}p{0.3\textwidth}X@{}} \toprule
            Type & Values and structure & Challenges & Features engineering\\ \midrule
            
            \textbf{Time-series}\newline
            \textit{Continuous}\newline 
            \textit{Regular}\newline
            \textit{Sporadic}
            &\begin{minipage}[t]{0.3\textwidth}{
            \begin{itemize}[leftmargin=*]  
                \item Time-stamped observations 
                \item Continuous, ordinal, categorical and/or multi-categorical
                \item Recorded continuously by medical devices, following a schedule by medical professional, or when necessary
            \end{itemize}}
            \end{minipage}
            &\begin{minipage}[t]{0.3\textwidth}{
            \begin{itemize}[leftmargin=*]  
                \item Observations are often \gls{mar} across time end dimensions, erroneous, or completely absent for certain patients.
                \item Time-series of different concepts are often highly correlated and their influence on one another must be accounted for.
             \end{itemize}}
             \end{minipage}
            & Imputation coupled with training \newline Regular \newline Data imputation \newline Binning in into fixed-size intervals \newline Combination of binning and imputation \\
            
            \textbf{Point-processes} 
            &\begin{minipage}[t]{0.3\textwidth}{
            \begin{itemize}[leftmargin=*]  
                \item Series of timestamped observations of one variable or medical concept per patient
            \end{itemize}}
            \end{minipage}
            &\begin{minipage}[t]{0.3\textwidth}{
            \begin{itemize}[leftmargin=*]  
                \item \todo{Intensity functions, paramtric models}
            \end{itemize}}
            \end{minipage}
            & Series of events reduced to the time interval between each consecutive occurrence. \\ 
            
            \textbf{Ordered \linebreak sequences} 
            & \begin{minipage}[t]{0.3\textwidth}{
            \begin{itemize}[leftmargin=*]  
                \item Ordered vectors representing one or more patients visits
                \item Medical codes associated with the diagnoses, procedures, measurements and interventions
            \end{itemize}}
            \end{minipage}
            & Variable length\newline High-dimensional\todo\newline Long-tail distribution of codes 
            & Sequences are projected into a trained embedding that preserves semantic meaning according to methods borrowed from NLP\\
            
            \textbf{Tabular}\newline Denormalized\newline Relational
            &\begin{minipage}[t]{0.3\textwidth}{
            \begin{itemize}[leftmargin=*]  
                \item Medical and demographic variables aggregated in tabular format
                \item Continuous, ordinal, categorical and/or multi-categorical features
            \end{itemize}}
            \end{minipage}
            & Medical history is aggregated into a fixed-size vector of binary or aggregated counts of occurrences and combined with demographic features.\\
            
            \bottomrule
        \end{tabularx}
    \end{table}

    \subsection{Data oriented GAN development}\label{subsec:data_gan_dev}

        \subsubsection{Auto-encoders and categorical features}\label{subsubsec:categorical}

            In what is to the best of our knowledge, the first attempt at developing a \gls{gan} for OHD. \citeauthor{Choi2017-nt} focus on the problem posed by the incompatibility of categorical and ordinal features with back-propagation. Their solution is to pretrain an \gls{ae} to project the samples to and from a continuous latent space representation. They keep the decoder portion along with its trained weights to form a component of \gls{medgan} \cite{Choi2017-nt}. It is incorporated into the generator and maps the randomly sampled input vectors from the real-valued latent space representation back to discrete features. This first exemplar of synthetic OHD generated by \gls{gan} inspires a series of enhancements.\par
            
            Early efforts were to improve the performance of \gls{medgan}. Among the first, \citeauthor{Camino2018-re} developed \gls{mc-medgan} changing the \gls{ae} component by splitting its output into a Gumbel-Softmax \cite{jang2016categorical} activation layer for each categorical variable and concatenating the results. \cite{Camino2018-re}. The authors also developed an adaptation based on recent training techniques: \gls{wgan} \cite{arjovsky2017wasserstein} and a \gls{wgan} (Briefed in \autoref{pan:wasserstein}) with Gradient Penalty \cite{gulrajani2017improved}. \gls{mc-wgan-gp} is the equivalent of \gls{mc-medgan} but with Softmax layers. The authors report that the choice of a model will depend on data characteristics, particularly sparsity.\par

\begin{figure}
    \footnotesize
\noindent
\tcbsetforeverylayer{autoparskip}
\tcbset{enhanced, nobeforeafter, width=1\linewidth}
\begin{tcolorbox}[arc=0.5mm, 
    colback=MidnightBlue!10!white, 
    coltext=MidnightBlue!90!black,  
    colframe=MidnightBlue!90!black,
    colbacktitle=MidnightBlue!80,
    leftrule=0mm,
    rightrule=0mm, 
    toprule=0mm, 
    bottomrule=0mm,
    box align=top,
    title={\begin{panel}Wasserstein's distance \label{pan:wasserstein}\end{panel}}]

In brief, the Wasserstein distance is a measure between two \glspl{pd} that has the property of always providing a smooth gradient. As the loss function of the discriminator, this property improves training stability and mitigates mode collapse. To make the equation tractable a 1-Lipschitz constraint must be introduced, creating another problem. In the words of the author: \begin{quote}
    "Weight clipping is a clearly terrible way to enforce a Lipschitz constraint. If the clipping parameter is large, then it can take a long time for any weights to reach their limit, [...] If the clipping is small, this can easily lead to vanishing gradients [...] However, we do leave the topic of enforcing Lipschitz constraints in a neural network setting for further investigation, and we actively encourage interested researchers to improve on this method." \cite{arjovsky2017wasserstein}
\end{quote} 
Sometimes this prevented the network from modelling the optimal function, but Gradient penalty, a less restrictive regularization replaced the clipping. \cite{Petzka2018}.

\end{tcolorbox}
\normalsize

\end{figure}

            Subsequent authors owing to the propensity of OHD to induce mode collapse widely adopted Wasserstein’s distance. Baowaly et al. developed \gls{medwgan} also based on \gls{wgan}, and \gls{medbgan} borrowing from Boundary-seeking \gls{gan} (BGAN) \cite{hjelm2017boundaryseeking} which pushes the generator to produce samples that lie on the decision boundary of the discriminator, expanding the search space. Both led to improved data quality, in particular \gls{medbgan} \cite{baowaly_2019_IEEE,baowaly_2019_jamia}. \citeauthor{Jackson_2019} tested \gls{medgan} on an extended dataset containing demographic and health system usage information, obtaining results similar to the original \cite{Jackson_2019}. \gls{healthgan}, based on \gls{wgan-gp}, includes a data transformation method adapted from the Synthetic Data Vault \cite{Patki_2016} to map categorical features to and from the unit numerical range \cite{Yale_2020}. 
        
        \subsubsection{Forgoing the autoencoder and introducing conditional training}\label{noauto}

            Claiming that the use of an \gls{ae} introduces noise, with \gls{emr-wgan}, \citeauthor{Zhang2020} dispose of the \gls{ae} component of previous algorithms and introduce a conditional training method, along with conditioned \gls{bn} and \gls{ln} techniques to stabilise training \cite{Zhang2020}. The algorithm was further adapted by \citeauthor{yan2020generating} as \gls{heterogan} to better account for the conditional distributions between multiple data types and enforce record-wise consistency. A recognized problem with \gls{medgan} was that it produced common-sense inconsistencies, such as gender mismatches in medical codes \cite{yan2020generating, Choi2017-nt}. \gls{heterogan} enforces constraints by adding specific penalties to the loss function, such as limit ranges for numerical categorical pairs and mutual exclusivity for pairs of binary features \cite{yan2020generating}. The algorithm also performs well on regular time-series of sleep patterns \cite{dash2019synthetic} \par

            To develop \gls{ctgan}, \citeauthor{Xu2019-ay} presume that tabular data poses a challenge to \gls{gan} owing to the non-Gaussian multi-modal distribution of real-valued columns and imbalanced discrete columns \cite{Xu2019-ay}. The fully connected layers, have adaptations to deal with both real-valued and categorical features. For real-valued features, it use mode-specific normalization to capture the multiplicity of modes. For discrete features, they introduce conditional training-by sampling to re-sample discrete attributes evenly during training, while recovering the real distribution when generating data.\par
            
            In other efforts, \citeauthor{torfi2019generating} develop \gls{corgan}, with a \gls{1d-cae} to capture neighboring feature correlations of the input vectors \cite{torfi2019generating}. \citeauthor{chincheong2020generation} use a \gls{ffn} based on Wasserstein's distance to evaluate the capacity of \glspl{gan} to model heterogeneous data of dense and sparse medical features \cite{chincheong2020generation}. \citeauthor{ozyigit2020generation} use the same approach, focusing on reproducing statistical properties \cite{ozyigit2020generation}.
            
        \subsubsection{Time-series}
            
             \citeauthor{esteban2017real} devise the LSTM-based \gls{rgan} and \gls{rcgan} to generate a regular time-series of physiological measurements from bedside monitors \cite{esteban2017real}. Curiously, the authors dismiss Wasserstein's distance explicitly, and generated each dimension of their time-series independently, where one would assume they are correlated. They observe a considerable loss of accuracy  on their utility metric. 

    \subsection{Task oriented GAN development}
        \subsubsection{Semi-supervised learning}

\begin{figure}
    \footnotesize
\noindent
\tcbsetforeverylayer{autoparskip}
\tcbset{enhanced, nobeforeafter, width=1\linewidth}
\begin{tcolorbox}[sidebyside,arc=0.5mm, 
    colback=MidnightBlue!10!white, 
    coltext=MidnightBlue!90!black,  
    colframe=MidnightBlue!90!black,
    colbacktitle=MidnightBlue!80,
    leftrule=0mm,
    rightrule=0mm, 
    toprule=0mm, 
    bottomrule=0mm,
    box align=top,
    title={\begin{panel}Transitional distribution \label{pan:transitional}\end{panel}}]

The \gls{ehrgan} generator is trained to decode a random vector $z$ mixed with the latent space representation of a real patient $h$ to produce a synthetic sample $\tilde{x}$ \cite{Che_2017}. A standard autoencoder (left) is trained to encode a real patient $x$ to and from a latent representation $h$, minimizing the reconstruction error with $\Bar{x}$. The decoder portion (left) is then trained to produce realistic synthetic samples $\tilde{x}$ from a combination of the random latent vector $z$ and the latent space encoding of a real patient $x$. The generator thus learns a transition distribution $p(\tilde{x}|x)$ with $x \thicksim p_{data}(x)$. The amount of contribution of the real sample is controlled by a random mask according to $\tilde{h} =m * z + (1 - m) \cdot h$. This method inspired from Variational Contrastive Divergeance prevents mode collapse by design and  learns an information rich transition distribution $p(\tilde{x}|x)$ around real samples $x$.
\tcblower
\includegraphics[width=1\columnwidth]{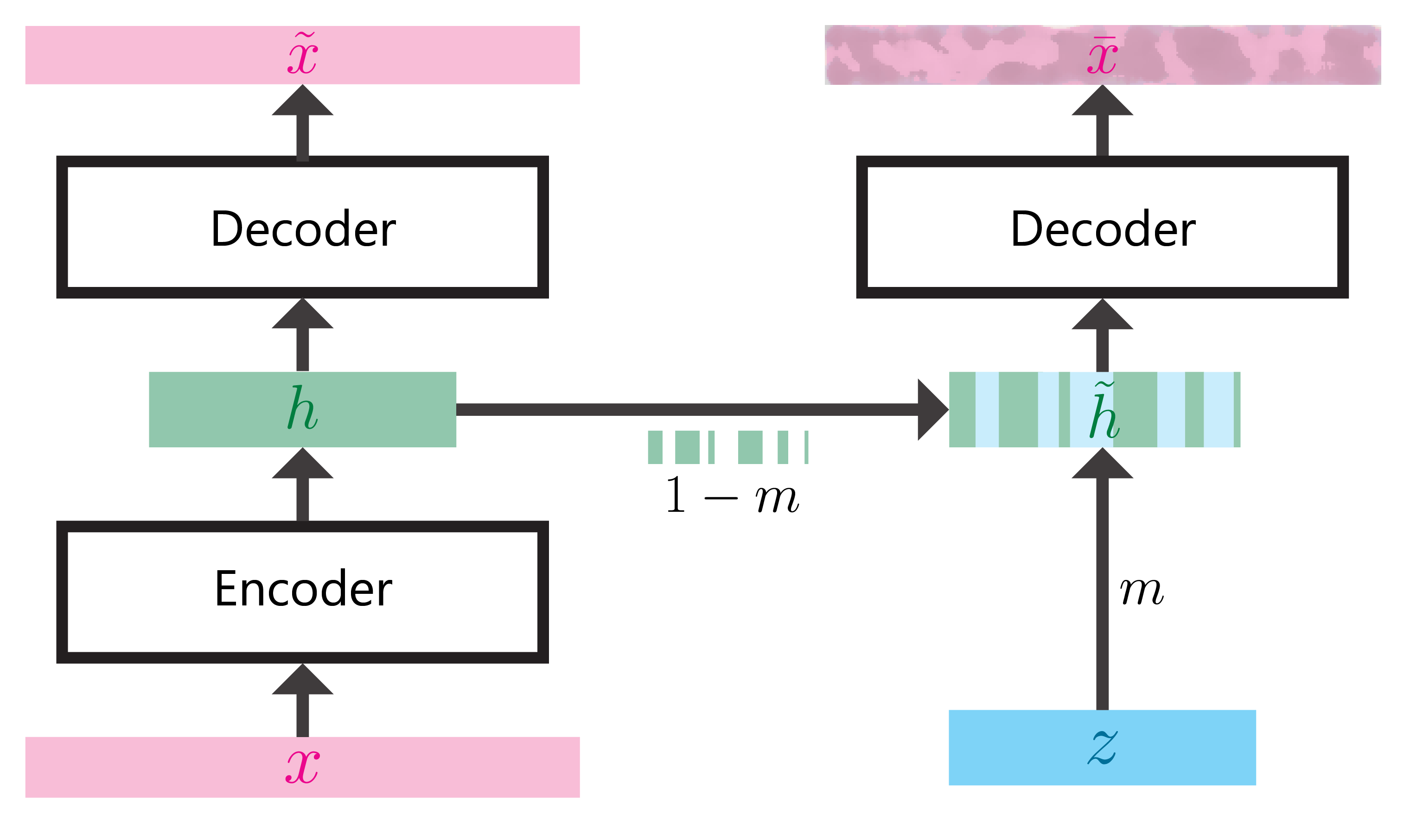}
\end{tcolorbox}
\normalsize
\end{figure}

            \gls{ehrgan} is developed for sequences of medical codes \citeauthor{Che_2017}. It learns a transitional distribution, combining an Encoder-Decoder \gls{cnn} \cite{Rankin2020} with \gls{vcd} \cite{Che_2017}. The \gls{ehrgan} generator is trained to decode a random vector mixed with the latent space representation of a real patient (See Panel \ref{pan:transitional}). The trained \gls{ehrgan} model is then incorporated into the loss function of a predictor where it can help generalization by producing neighbors for each input sample.\par
            
            \Gls{ssl} is commonly used to augment the minority class in imbalanced datasets, with techniques such as \gls{st} and \gls{ct}. \citeauthor{yang2018unpaired} improves on both by incorporating a \gls{gan} in the procedure \cite{yang2018unpaired}. The \gls{gan} is first trained on the labelled set and used to re-balance it. A prediction task with a classifier ensemble is then executed and the data points with highest prediction confidence are labelled. The process is iterated until labelling expansion ceases. As a final step, the \gls{gan} is trained on the expanded labelled set to generate an equal amount of augmentation data. The authors obtained improved performance in a number of classification tasks and multiple tabular datasets.
    
    \subsubsection{Domain translation}
    
        To address the heterogeneity of healthcare data originating from different sources, \citeauthor{Yoon2018-radial} combines the concepts of cycle-consistent domain translation from \gls{cycle-gan} \cite{Zhu_2017} and  multi-domain translation from Star-GAN \cite{choi2017stargan} to build \gls{radialgan} to translate heterogeneous patient information from different hospitals, correcting features and distribution mismatches \cite{Yoon2018-radial}. One encoder-decoder pair per data endpoint are trained to map records to and from a shared latent representation for their respective endpoint. 
    
    \subsubsection{Individualized treatment effects}
    
        The task of estimating \glspl{ite} is an ongoing problem. \glspl{ite} refer to the response of a patient to a certain treatment given a set of characterizing features. The problem is that counterfactual outcomes are never observed or treatment selection is highly biased \cite{Yoon2018-ite, mcdermott2018semi, walsh2020generating}. In \gls{ganite} \citeauthor{Yoon2018-ite} propose a solution by using a pair of \glspl{gan}: one for counterfactual imputation and another for \gls{ite} estimation \cite{Yoon2018-ite}. The former captures the uncertainty in unobserved outcomes by generating a variety of counterfactuals. The output is fed to the latter, which estimates treatment effects and provides confidence intervals.\par
    
        \citeauthor{mcdermott2018semi} developed \gls{cwr-gan} to leverage large amounts of unpaired pre/post-treatment time-series in \gls{icu} data for the estimation of \glspl{ite} on physiological time-series \cite{mcdermott2018semi}. \gls{cwr-gan} is a joint regression-adversarial  \gls{ssl} approach inspired by \gls{cycle-gan}. The algorithm has the ability to learn from unpaired samples, with very few paired samples, to reversibly translate the pre/post-treatment physiological series.\par 
    
        \citeauthor{chu2019treatment} address the problem of data scarcity by designing \gls{adtep}. The algorithm can harness the large volume of \gls{ehr} data formed by triples of non-task specific patient features, treatment interventions and treatment outcomes \cite{chu2019treatment}. \gls{adtep} learns representation and discriminatory features of the patient, and treatment data by training an \gls{ae} for each pair of features. In addition to \gls{ae} reconstruction loss, a second discriminator is tasked with identifying fake treatment feature reconstructions. Finally, a fourth loss metric is calculated by feeding the concatenated latent representations of both \glspl{ae} to a \gls{lr} model aimed at predicting the treatment outcome \cite{chu2019treatment}.\par
    
        Like \citeauthor{esteban2017real}, \citeauthor{Wang_2019} demonstrated an algorithm to generate a time series of patient states and medication dosages pairs using \gls{lstm}. In contrast to \gls{rgan} and \gls{rcgan}, in \gls{sc-gan}, patients state at the current time-step informs the concurrent medication dosage, which in turn affects the patient state in the upcoming time-step \cite{Wang_2019}. \gls{sc-gan} overcame a number of baselines on both statistical and utility metrics.
    
    \subsubsection{Data imputation and augmentation}
    
        \gls{gan} are naturally suited for data imputation, and can mitigate missingness. Statistical models developed for the multiple imputation problem increase quadratically in complexity with the number of features, while the expressiveness of deep neural networks can efficiently model all features with missing values simultaneously.\par
        In that regard, \citeauthor{yoon2018imputation} adapted the standard \gls{gan} to perform imputation on real-valued features \gls{mar} in tabular datasets \cite{yoon2018imputation}. In \gls{gain}, the discriminator must classify individual variables as real or fake (imputed), as opposed to the whole ensemble. Additional input, or hint, containing the probability of each component being real or imputed is fed to the discriminator to resolve the multiplicity of optimal distributions that the generator could reproduce. The model performs considerably better than five state-of-the-art benchmarks. \gls{gain} was later adapted by \citeauthor{Yang_2019_impute_ehr} to also handle categorical features using fuzzy binary encoding, the same technique employed in \gls{healthgan}. In parallel, \citeauthor{Camino2019} apply the same \gls{vs} technique they used fir \gls{medgan} to adapt \gls{gain} and run a benchmark against different types of \gls{vae}.\par
    
        The distribution estimated by a generator can compensate for lack of diversity in a real sample, essentially filling in the blanks in a manner comparable to data imputation. In such cases, data sampled from this distribution has the potential to help improve generalization in training predictive models. As an example, we mentioned generating unobserved counterfactual outcomes \cite{yoon2018imputation}, and generating neighboring samples to help generalization in predictors \cite{Che_2017}.\par 
        The adversarially trained \gls{rmb} developed by \citeauthor{Fisher2019} enabled them to simulate individualized patient trajectories based on their base state characteristics. Due to the stochastic nature of the algorithm, generating a large number of trajectories for a single patient can provide new insights on the influence of starting conditions on disease progression or quantify risk \cite{Fisher2019}.
        
    \subsection{Model validation and data evaluation}
    
        To assess the solution to a generative modelling problem, it is necessary to validate the model, and to verify its output. \gls{gan} aim to approximate a data distribution $P$, using a parameterized model distribution $Q$ \cite{Borji2018-fy}. Thus, in evaluating the model, the goal is to validate that the learning process has led to a sufficiently close approximation. What this means in practice is hard to define. The concept of "realism" finds more natural application to images of text, but becomes ambiguous when faced with the complexity of health data.\par
        
        \citeauthor{walsh2020generating} employ the term "statistical indistinguishability" and define it as the inability of a classification algorithm to differentiate real from synthetic samples \cite{walsh2020generating}. The terms covers almost all evaluation methods employed in the publications, which can be divided into two broad categories: those aimed at evaluating the statistical properties of the data directly, and those aimed at doing so indirectly by quantifying the work that can be done with the data. There are, nonetheless a few attempts of a qualitative nature, more in line with the concept of realism. 

        \subsubsection{Qualitative evaluation}
        
            Visual inspection of projections of the \gls{sd} is a common theme, serving mostly as a basic sanity check, but occasionally presented as evidence. The formal qualitative evaluation approaches found in the literature are mainly Preference Judgement, Discrimination Tasks or Clinician Evaluation and are generally carried out by medical professionals (Borji 2018).
                \begin{itemize}
                    \item \textbf{Preference judgment} The task is choosing the most realistic of two data points in pairs of one real and one synthetic \cite{Choi2017-nt}.
                    \item \textbf{Discrimination Tasks} Data points are shown one by one and must be classified as real or synthetic \cite{Beaulieu-Jones2019-ct}.
                    \item \textbf{Clinician Evaluation} Rather than classifying the data points, they must be rated for realism according to a predefined numerical scale. \cite{Beaulieu-Jones2019-ct}. Significance is determined with a statistical test such as Mann-Whitney.
                    \item \textbf{Visualized embeding} The real and synthetic data samples are plotted on a graph or projected into an embeding such as \gls{t-sne} or PCA and compared visually. \cite{cui2019conan, yu2019rare, zhu_2020, yale2019ESANN, Yang_2019_ehr,Beaulieu-Jones2019-ct, tanti2019, dash2019synthetic}.
                    \item \textbf{Feature analysis} In certain fields, the data can be projected to representations that highlight patterns or properties that can be easily visually assessed. While this does not provide conclusive evidence of data realism, it can help get a better understanding of model behaviour during training. As an example, typical and easily distinguishable patterns in EEG and ECG bio-signals. \cite{Harada2019}
                \end{itemize}
    
        In general, qualitative evaluation methods based on visual inspection are weak indicators of data quality. At the dataset or sample level, quantitative metrics provide more convincing evidence of data quality (Borji 2018).
        
        \subsubsection{Quantitative evaluation}
        
            Quantitative evaluation metrics can be categorized into three loosely defined groups: comparing the distributions of real and synthetic data as a whole, assessing the marginal and conditional distributions of features, and evaluating the quality of the data indirectly by quantifying the amount of work that can be done with the data, referred to as utility.
            
            \begin{itemize}
                \item \textbf{Dataset distributions}
                A summary of metrics is presented in Tab. \ref{tab:3:distributions}.
                \item \textbf{Feature Distributions}
                If the model has learned a realistic representation of the real data it should produce \gls{sd} that possesses the same quantity and type of information content. Authors attempt by various metrics to determine if the statistical properties of the \gls{sd} agree with those of the real data. These metrics are presented in Table \ref{tab:3:statistics}. Although statistical similarity provides strong support for the behavior of the learning process, it is not necessarily informative about their validity. They are often ambiguous and can be found to be misleading upon further investigation. Given the complexity of health data, low level relations are unlikely to paint a full picture. Authors often state that no single metric taken on its own was sufficient, and that a combination of them allowed deeper understanding of the data.
                \item \textbf{Data utility}
                 Utility-based metrics, presented in Table \ref{tab:3:augmentation}, often provide a more convincing indicator of data realism. On the other hand, they mostly lack the interpretability of statistical metrics. We took the liberty of placing these into one of two categories: tasks mostly defined only for evaluation (Ad hoc utility metrics) or tasks based on real-world applications (Application utility metrics). Note that this distinction is not based on a rigorous definition, but serves to facilitate comparison.
                \item \textbf{Analytical} The analytical methods were mainly employed for evaluation, but can also provide a better understanding of the and its behavior.
                \begin{itemize}
                    \item \textsl{Feature Importance} The important features (\gls{rf}) and model coefficients (\gls{lr}, \gls{svm}) of predictors. \cite{esteban2017real,Xu2019-ay,Yoon2020-anon,chin2019generation, Beaulieu-Jones2019-ct}.
                    \item \textsl{Ablation study}  The performance of the model is compared against impaired version. This helps determining if the novel component of the algorithm contributes significantly to performance \cite{cui2019conan, Che_2017, mcdermott2018semi, Yoon2018-radial, chincheong2020generation}.
                \end{itemize}
            \end{itemize}
            
                 \begin{table}[H]         
 \footnotesize  
 \setlength{\extrarowheight}{0.5em}
 \caption{Metrics employed to validate trained models based on the comparison of distributions.\label{tab:3:distributions}}              
    \begin{tabular}{@{} p{0.2\textwidth} p{0.8\textwidth} @{}}\toprule                          
    Metric & Description \\ \midrule                                  
    
    \gls{kld} 
    & Non-symmetric measure of difference between two \glspl{pd}, related to relative entropy. Given a feature $X$, $p(x)$ and $q(x)$ the \gls{pd} of the real and synthetic data respectively, the \gls{kld} of $q(x)$ from $p(x)$ is the amount of information lost when $q(x)$ is trained to estimate $p(x)$ \cite{klb2008, Goncalves2020}. \\       
    
    \gls{rdp} 
    & Alternative measure of divergence, which includes \gls{kld} as a special case. The \gls{rdp} includes a parameter $\alpha$ that gives it an extra degree of freedom, becoming equivalent to the Shannon-Jensen divergence when $\alpha \longrightarrow 1$. It showed a number of advantages when compared to the original \gls{gan} loss function, and removed the need for gradient penalty \cite{VanBalveren2018, tanti2019}\\ 
    
    Jaccard similarity & Measure of similarity and diversity defined on sets as the size of the intersection over the size of the union \cite{ozyigit2020generation, Yang_2019_ehr, Wikipediacontributors}.\\

    2-sample test (2-ST) 
    & Statistical test of the null hypotheses the real and \gls{sd} samples came from the same distribution. and synthetic, originate from the same distribution through the use of a statistical test such as \gls{ks} or \gls{mmd}.\cite{Fisher2019,baowaly_2019_IEEE,baowaly_2019_jamia,esteban2017real}\\     
    
    Distribution of Reconstruction Error 
    & Compares the distributions of reconstruction error for the \gls{sd} and the training set versus the \gls{sd} and a held out testing set. Calculated according to the Nearest-neighbor metric or other measures of distance. A significant difference would indicate over-fitting and can evaluated with a statistical test, such as \gls{ks}. \cite{esteban2017real}\\
    
    Latent space projections 
    & Real and synthetic samples are projected back into the latent space, or encoded with a \gls{beta-vae}, comparing the dimensional mean of the variance or the distance between mode peaks \cite{Zhang2020}. See Section \ref{sec:latent-space} for examples of how the latent space encoding can interpreted. \\
    
    \glspl{dsm} 
    & Comparison of the \gls{pd} with \glspl{dsm}. For instance the Quantile-Quantile (Q-Q) plot for point-processes \cite{Xiao2017-lh}. See Section \ref{sec:evaluation-cqm} for a notion of how \glspl{dsm} could apply to \gls{ehr} data.\\                              
    Classifier accuracy &  
    Accuracy of a classifier trained to discriminate real from synthetic units. Predictor accuracy around 0.5 would indicate indistinguishability. \cite{Fisher2019,walsh2020generating}\\       
    
    \bottomrule                      
    \end{tabular}         
\end{table}
                \begin{table}[H]
        \footnotesize
        \setlength{\extrarowheight}{0.5em}
        \caption{Metrics based on evaluating the statistical properties of the synthetic data distribution. \label{tab:3:statistics}}
        \begin{tabularx}{\textwidth}{@{} p{0.3\textwidth} X @{}}\toprule
            Metric & Description\\ \midrule
            
            Dimensions-wise distribution & 
            The real and synthetic data are compared feature-wise according to a variety of methods For example, the Bernoulli success probability for binary features, or the Student T-test for continuous variables, and Pearson Chi-square test for binary variables is used to determine statistical significance \cite{Beaulieu-Jones2019-ct,Choi2017-nt,chin2019generation,yan2020generating,baowaly_2019_IEEE,baowaly_2019_jamia,ozyigit2020generation,tanti2019, Yoon2020-anon, tanti2019, Fisher2019, Che_2017, Wang_2019, yale2019ESANN, chincheong2020generation, ozyigit2020generation}.\\
            
            Inter-dimensional correlation & 
            Dimension-wise Pearson coefficient correlation matrices for both real and synthetic data \cite{Beaulieu-Jones2019-ct, Goncalves2020, torfi2019generating,Frid_Adar_2018,ozyigit2020generation, Yang_2019_ehr, Yoon2020-anon, zhu_2020, Yoon2020-anon, walsh2020generating, yale2019ESANN, ozyigit2020generation, dash2019synthetic, Bae2020}.\\
           
            Cross-type Conditional Distribution & 
            Correlations between categorical and continuous features, comparing the mean and standard deviation of each conditional distribution \cite{yan2020generating}.\\
            
            Time-lagged correlations & 
            Measures the correlation between features over time intervals.
            \cite{Fisher2019,walsh2020generating}.\\
            
            Pairwise mutual information & 
            Checks for the presence multivariate relationships pair-wise for each feature, as a measure of mutual dependence \cite{Rankin2020}. Quantifies the amount of information obtained about a feature from observing another.\\
            
            First-order proximity metric & 
            Defined over graphs, captures the direct neighbor relationships of vertices. \citeauthor{Zhang2020} applied to graphs built from the co-occurrence of medical codes and compared the results between real and synthetic data \cite{Zhang2020}.\\
            
            Log-cluster metric & 
            Clustering is applied to the real and synthetic data combined. The metric is calculated from the number of real and synthetic samples that fall in the same clusters \cite{Goncalves2020}.\\
            
            Support coverage metric & 
            Measures how much of the variables support in the real data is covered in the synthetic data. Support is defined as the percentage of values found in the synthetic data, while coverage is the reverse operation. The metric is calculated as the average of the ratios over all features. Penalizes less frequent categories that are underrepresented \cite{Goncalves2020}.\\
 
            Proportion of valid samples & 
            Defined by \citeauthor{Yang_2019_ehr} as a requirement for records to contain both disease and medication instances. \cite{Yang_2019_ehr}.\\
            
            \gls{pca} Distributional Wassertein distance 
            & The Wassertein distance is calculated over k-dimensional \gls{pca} projections of the real and synthetic data \cite{tanti2019}.\\
            
            \bottomrule
        \end{tabularx}
    \end{table}
                \begin{table}[H]
        \footnotesize
         \setlength{\extrarowheight}{0.5em}
        \caption{Metrics based on evaluating the utility of the synthetic data on practical tasks.}\label{tab:3:augmentation}
        
        \begin{tabularx}{\textwidth}{@{} p{0.2\textwidth} X @{}} \toprule
        Metric & Description \\ \midrule
        
        \multicolumn{2}{c}{\textbf{Data utility metrics}}\\ \midrule
        
        \gls{dwp} & Each variable is in turn chosen as the prediction target label and the remaining as features. Two predictors are trained to predict the label, one from the synthetic data and another from a portion of the real data. Their performance is compared on the left out real data \cite{Choi2017-nt,Camino2018-re,Goncalves2020,yan2020generating, tanti2019, baowaly_2019_IEEE}.\\
        
        \gls{arm} & \gls{arm} aims to the discovery of relationships among a large set of variables, commonly occurring variable-value pairs \cite{Agrawal1993}. The rules obtained from the real and synthetic data are compared \cite{baowaly_2019_IEEE,baowaly_2019_jamia,BaeAnomiGAN2020,yan2020generating}.\\
        
        Training utility & Performance of predictors trained on the synthetic data, often in comparison with the real data or data generated with \gls{dp} \cite{BaeAnomiGAN2020}.\\
        
        \gls{trts} & Accuracy on real data of some form of predictor trained on synthetic data \cite{Beaulieu-Jones2019-ct, Rankin2020, Yoon2020-anon}. \\ 
        
        \gls{tstr} & Accuracy on synthetic data of some form of predictor trained on real data   \cite{BaeAnomiGAN2020, Yoon2020-anon, Jordon2019}.\\
        
        Discriminator & A predictor is trained to discriminate synthetic from real sample. An accuracy value of 0.5 would indicate that they are indistinguishable \cite{Fisher2019, walsh2020generating, yale:hal-02160496}.\\ 
        
        Siamese discriminator & A pair of identical \gls{ffn} each receive either a real sample or a synthetic sample. Their output is passed to a third network which outputs a measure of similarity \cite{torfi2019generating}.\\\midrule

        \multicolumn{2}{c}{\textbf{Applied utility metrics}}\\ \midrule
        
        Data augmentation & A predictor is trained on a combination dataset of real and synthetic data or real data with missing values imputed and performance is compared with the same predictor trained on real data alone \cite{Yoon2020-anon, Yang_2019_cdss, Yang_2019_ehr}.\\
        
        Model augmentation & The trained generative model is incorporated into a predictor's activation function by generating an ensemble of proximate data points for each instance, thereby improving generalization \cite{Che_2017}.\\
        
        Accuracy & The prediction performance of the model is compared against benchmarks of the same type on real data \cite{cui2019conan, Yoon2018-ite, Che_2017, yu2019rare, zhu_2020, baowaly_2019_IEEE, Wang_2019, walsh2020generating, yoon2018imputation, mcdermott2018semi, Yang_2019_ehr, Yoon2018-radial, Xu2019-ay, Beaulieu-Jones2019-ct, BaeAnomiGAN2020}. Models trained to make forward predictions from past observations or from real data transformed with a known function can simply be evaluated for accuracy. For example, the \gls{rmse} on time-series \cite{Xiao2018-aj,mcdermott2018semi,yoon2018imputation,Yang_2019_cdss, zhu_2020}.\\
        
        \bottomrule
        
        \end{tabularx}
\end{table}

    \subsection{Alternative evaluation}
        In their publications, \citeauthor{Yale_2020} propose refreshing approaches to evaluating the utility of \gls{sd}. For example, they organized a hack-a-thon type challenge involving the data. During the event, students were tasked with creating classifiers, while provided only with \gls{sd} \cite{Yale_2020}. They were then scored on the accuracy of their model on real data.\par
        In more rigorous initiatives, they attempted (successfully) to recreate the experiments published in medical papers based on the MIMIC dataset using only data generated from their model \gls{healthgan}. In a subsequent version of their article, the authors evaluate the performance of their model against traditional privacy preservation methods by using the trained discriminator component of \gls{healthgan} to discriminate real from synthetic samples.
        
    \subsection{Privacy}
        Some authors conducted a privacy risk assessment to evaluate the risk of reidentification. The empirical analyses were based on the definitions of \gls{mi}, \gls{ad}  \cite{Choi2017-nt,Goncalves2020,yan2020generating,chen2019ganleaks, chincheong2020generation} and the \gls{rr} \cite{Zhang2020}. Cosine similarities between pairs of samples was also used \cite{torfi2019generating}. Most studies report low success rates for these types of attacks, and little effect from the sample size, although \citeauthor{chen2019ganleaks} note that sample sizes under 10k lead to higher risk. \citeauthor{Goncalves2020} evaluated \gls{mc-medgan} against multiple non-adversarial generative models in a variety of privacy compromising attacks, including \gls{ad}, obtaining inconsistent results for \gls{mc-medgan} \cite{Goncalves2020}. While this is not mentioned by the authors, multiple results reported in the publication point to the fact that the \gls{gan} was not properly trained or suffered mode-collapse.In black-box and white-box type attacks, including the LOGAN \cite{hayes2017logan} method, \gls{medgan} performed considerably better than \gls{wgan-gp} \cite{chen2019ganleaks}, the algorithm which served as basis for improvements to \gls{medgan} in publications discussed in Section \ref{subsubsec:categorical}. Overall, the author notes that releasing the full model poses a high risk of privacy breaches and that smaller training sets (under 10k) also lead to a higher risk. \par
        \subsubsection{The status of fully synthetic data in regards to current privacy regulations}
        
            It seems intuitively possible that the artificial nature of \gls{sd} essentially prevents associations with real patients, however the question is never directly addressed in the publications. An extensive Stanford Technological Review legal analysis of \gls{sd} concluded that laws and regulations should not treat \gls{sd} indiscriminately from traditional privacy preservation methods \cite{bellovin2019privacy}. They state that current privacy statutes either outweigh or downplay the potential for \gls{sd} to leak secrets by implicitly including it as the equivalent of anonymization. 
            
        \subsubsection{Traditional privacy}
        
            Numerous attempts at applying traditional privacy guarantees, such as deferentially-private stochastic gradient descent can also be found in other fields, as well as in healthcare \cite{Beaulieu-Jones2019-ct, esteban2017real,chincheong2020generation, BaeAnomiGAN2020}. By limiting the gradient amplitude at each step and adding random noise, AC-GAN could produce useful data with $\epsilon=3.5$ and $\delta<10^{-5}$ according to the definition of differential privacy. \par 
        
        \subsubsection{Moving forward safely}
        
            Some have put forward the notion that preventing over-fitting and preserving privacy may not be conflicting goals \cite{Wu2019-ui,Mukherjee2019-vu,Zhu2020-oj}. Letting go of the negative connotation, we can explore the benefits such as improving generalization, stabilizing learning and building fairer models \cite{Zhu2020-oj} and the use of \glspl{gan} to optimize the trade-off \cite{Chen2019-mh}.\par
            
            \begin{itemize}
                \item \citeauthor{BaeAnomiGAN2020} ensure privacy with a probabilistic scheme that ensure indistinguishably, but also maximizes utility. Specifically, a multiplicative perturbation by random orthogonal matrices with input entries of $k x m$ medical records and a second second discriminator in the form of a pre-trained predictor \cite{BaeAnomiGAN2020}.
                \item In privGAN \cite{Mukherjee2019-vu}, an adversary is introduced, forcing the generator to produce samples that minimize the risk of \gls{mi} attacks, in addition to cheating the discriminator. The combination of both goals has the explicit effect of preventing over-fitting, and their algorithm produces samples of similar quality to non-private \gls{gan}.
            \end{itemize}
           \subsubsection{Alternative views of privacy}
            The discordance between the theoretical concepts of DP, which are  based ultimately on infinite samples, and the often insufficient data on which the probability of disclosure is calculated remains deficient. Therefore, Yoon et al. have postulated an intriguing alternative view of privacy \cite{Yoon2020-anon}. They propose to emphasize measuring identifiability of finite patient data, rather than the probabilistic disclosure loss of DP based on unrealistic premises. Simplistically, they define identifiability as the minimum closest distance between any pair of synthetic and real samples. This echoes the concept of t-closeness \cite{Li2010-qq}. In their implementation, the generator receives both the usual random seed and a real sample as input. This has the effect of mitigating mode collapse, but also of raises the risk of reproducing the real samples. The discriminator is equipped with an additional loss metric based on a measure of similarity between the original sample and the generated one, thus ensuring a tune-able threshold of identifiability. Their results on a number of previously discussed evaluation metrics are encouraging.\par
            
            In a similar approach, \citeauthor{Yale_2020} broke away from the theoretical guarantees of traditional methods with a measure native to \glspl{gan}. Their proposal is a metric quantifying the loss of privacy, a concept more aligned with the objective of \gls{gan} to minimize the loss of data utility \cite{yale:hal-02160496,p2019}. They point out the advantage of concrete measurable values of loss in utility and privacy when making the decision of releasing sensitive data. Briefly, the Nearest Neighbor Adversarial Accuracy measures the loss in privacy based on the difference between two nearest neighbor metrics. The  first component is the proportion of synthetic samples that are closer to any real sample than any pair of real samples. The second component is the reverse operation. In a subsequent paper, \gls{healthgan} evaluated against traditional privacy preservation methods with a variant of the IA based on the nearest neighbor metric. \gls{healthgan} performs considerably better than all other methods, while still maintaining utility on a prediction task.

    \section{Discussion}

\subsection{Applications of GANs for health data and innovation}

Overall, the published \gls{gan} algorithms for \gls{ohd} provided equivalent or superior performance versus the statistical modeling-based methods against which they were benchmarked. Importantly, their capabilities are highly relevant to the medical field: domain translation for unlabeled data, conditional sampling of minority classes, data augmentation, learning from partially labeled or unlabeled data, data imputation, and forward simulation of patient profiles. While some of these claims are overoptimistic or lack convincing evidence, they paint an encouraging picture for the value of synthetic \gls{ohd} and the transformative effect it could have on healthcare initiatives and scientific progress.\par

The ongoing Covid-19 pandemic has brought unprecedented levels of cooperation between scientists from around the world. The urgency of obtaining data has highlighted on difficult terms the need for novel ways of sharing and generating data \cite{bandara_improving_2020, Cosgriff_2020}. Global concerted efforts were highly successful, but also required adaptation, with some proposing exemptions from the GDPR \cite{mclennan_covid-19_2020}. Data sharing was limited to aggregate counts, rather than at the patient level, limiting the depth of analyses. \par

In the beginning of an epidemic, the scarcity of data can be compensated with synthetic data. Building generative statistical methods in such conditions is a difficult task \cite{Latif2020-ol}. While additional data becomes available to fine-tune the model, so do the number of features and the complexity of the model. This was attempted by Synthea \cite{Walonoski_2017} in the early months of the pandemic, with humble results, nonetheless they were used in many online challenges, hackathons, and conferences. 
\begin{quote}
    The authors state that if one takes "[...] Field Marshall Moltke’s notion of “no plan survives contact with the enemy” as true and expands the scope to modeling and simulation, then we might say that “no model survives contact with reality.” \cite{walonoski_synthea_2020}. We would argue that \glspl{gan} grow stronger in contact with reality. 
\end{quote}
Generative models refine their representation as more data is provided and could be combined with current methods of forecasting. When the amount of ground truth data is small, semi-supervised learning simulations can improve the performance of predictors \cite{dahmen_synsys_2019}. Domain translation, as demonstrated in \gls{radialgan}, would be exceptionally useful to combine datasets from disparate localities. In a recent publication, two different data augmentation techniques provided a significant increase in sensitivity and specificity for the detection of COVID-19 infections, one of which producing \gls{sd} with a \gls{gan} \cite{Sedik2020-tx}.

\subsection{Challenges posed by OHD}

The challenges posed by health data for \glspl{gan} are obvious, a number of recurrent factors influence the outcome of efforts to develop them. These problems are not limited to generative algorithms, but also \gls{ml} in general. For generative models, multi-modality caused the most trouble in achieving a stable training procedure. At the outset, preventing \gls{mode-collapse} attracted the most research efforts, in addition to data combinations of categorical and real-valued features. A rapid succession of efforts aimed at improving \gls{medgan} by incorporating the latest machine learning techniques showed continued improvements. However, taken as a whole the efforts were haphazard in their methods and metrics. Often yielding unsurprising results, considering the techniques were known to improve performance across a broad range of applications. This is expected in a new field of application, and more concerted efforts to systematically approach the problems should progressively form.\par

\subsubsection{Feature engineering}
We observed that majority of methods included in the review made use of heavily transformed representations of patient records. This is in part due to the inconvenient properties of health data, such as missingness. However, it is somewhat apparent that the main motive is to accommodate existing algorithms. Along with demographic variables, \gls{ohd} data mostly takes the form of triples composed by (1) a timestamp, (2) a medical concept and (3) the recorded value. Their count is different for each patient, irregular intervals between each triple and the number of possible values in a dimensions can be huge. Moreover, there are generally multiple episodes of care, each with a different cause. These properties are not typically considered practical for machine learning. \par

At varying degrees, depending on the transformations, information is being lost or bias is introduced. For example, when data are reduced by aggregation to one-hot encoding, the complex relationships found in medical data are, for the most part eliminated. Similarly, information is lost when forcing real-valued time-series into a regular representation, by truncating, padding, binning or imputation. Moreover, it is highly unlikely that the data is missing at random, introducing the potential for bias when a large part of the real data is rejected on this basis, or the medical codes are truncated to their parent generalizations \cite{Zhang2020, Choi2017-nt}. 

\subsection{From innovation to adoption: Evaluation metrics and benchmarking}
Interesting innovations were demonstrated, and progress has good momentum. Their application and adoption will undoubtedly be more sluggish, as has been the case with predictive \gls{ml}. For good reason, the bar is set high in demonstrating consistent outcomes and ensuring patient safety. While the problem of \gls{mode-collapse} has been alleviated, evidence has yet to be provided with regards to ensuring that the finer details of the distribution are estimated with sufficient granularity to produce realistic patient profiles. Consistent behavior and reproducible results will be required to expect any significant adoption. In regards to evaluation, it is manifest that the choice of optimal metrics and indicators is still being explored. The fact is that the efforts are far from consistent or systematic. As an example, competing methods are often compared with different metrics or with contradictory results in different datasets \cite{baowaly_2019_IEEE,baowaly_2019_jamia,Camino2018-re,Choi2017-nt,Zhang2020}. Overall, none of the evaluation metrics addressed the concept of realism in synthetic data.\par

\subsubsection{Qualitative realism}
Qualitative evaluation, in its current form, provides little evidence. For medical experts, these representations are meaningless. As such, the results of qualitative evaluation often state that synthetic data is indistinguishable from the real data \cite{Choi2017-nt,Wang_2019}. It is doubtful that they could in fact be distinguished. \citeauthor{esteban2017real} found that participants avoided the median score and were not confident enough to choose either extreme \cite{esteban2017real}.\par

\begin{figure}
    \footnotesize
\noindent
\tcbsetforeverylayer{autoparskip}
\tcbset{enhanced, nobeforeafter, width=1\linewidth}
\begin{tcolorbox}[sidebyside,arc=0.5mm, 
    colback=MidnightBlue!10!white, 
    coltext=MidnightBlue!90!black,  
    colframe=MidnightBlue!90!black,
    colbacktitle=MidnightBlue!80,
    leftrule=0mm,
    rightrule=0mm, 
    toprule=0mm, 
    bottomrule=0mm,
    box align=top,
    title={\begin{panel}Representation and visualisation \label{pan:visualisation}\end{panel}}]

\citeauthor{Ledesma2016-hn} describe the problem of medical data representation and visualization learnedly, from information quality and usefulness, timescales and perception, to user satisfaction and aesthetics. The evaluation of their solution is extensive, detailed and rigorous, done according to the well known Nielsen's heuristics for Human-Computer Interaction \cite{nielsend}.  Interested readers can find the remainder here \href{https://www.nngroup.com/articles/ten-usability-heuristics/}{Ten Usability Heuristics.} While this may seem like total digression towards graphic design, it is rather to illustrate the complexity of aspects to be considered before representing data in a evaluation task.
\tcblower

\textbf{Principle \#2: Match between system and the real world}:\\ \textit{The system should speak the users' language, with words, phrases and concepts familiar to the user, rather than system-oriented terms. Follow real-world conventions, making information appear in a natural and logical order. }
\\
---
\\
\textbf{Principle \#4 Consistency and standards}\\
\textit{Users should not have to wonder whether different words, situations, or actions mean the same thing. Follow platform conventions.
}

\end{tcolorbox}
\normalsize
\end{figure}%

In their evaluation of \gls{medgan}, \cite{yale:hal-02160496} argue that the positive resemblance of plotted feature distributions is due to the fact that the model's architecture tends to favor reproducing the means and probabilities of each diagnosis column. They note that synthetic data contains samples with an unusually high number of codes, which is not apparent in the plots. Their hypothesis is that these samples are used by the algorithm to discharge the rare medical codes with weak correlation, in an effort to balance the distributions. However, they stated in their experiments that comparing \gls{pca} plots of real and synthetic data  was nonetheless insightful to get an impression of their behavior \cite{Yale_2020} If visual inspection is to be used, it should be done systematically according established frameworks (See Panel \ref{pan:visualisation})\par

\subsubsection{Quantitative fitness}
Reproducing aggregate statistical properties is rather unconvincing evidence that a model has learned to reproduce the complexity of patient health trajectories. In some cases the statistical metrics may be contradictory, such as when the ranking of medical frequencies in the data are wrong, but augmentation leads to improved performance \cite{Che_2017}. \citeauthor{Choi2017-nt} found that although the synthetic sample seemed statistically sound, it contained gross errors such as gender code mismatches and suggested the use of domain-specific heuristics \cite{Choi2017-nt}. \gls{heterogan} was an encouraging step in this direction, but do not represent a solution. Conditional training methods have led to improvements. For example, when labels corresponding to sub-populations or classes are used to condition the generative process. \citeauthor{Zhang2020} showed that conditioned training with categorical labels, in this case age ranges, improves utility for small datasets \cite{Zhang2020}. \par

Utility-based metrics do overall provide a more solid evaluation of data quality. However, they only confirm the value of the data according to a narrow context. They are indicative of realism so far as a patient's state is indicative of a medical outcome. Moreover, they do not provide any insight about the validity of the relations found in a patient record and its overall consistency. While such consideration was found in sparingly in the publications, extensive research available on the subject of medical information representation. The complexity of health data and its variety make it a considerable, but captivating challenge.\par

\subsubsection{Constraints}
As described in Section \ref{noauto}, \gls{heterogan} introduces constraint-based loss. Based on the distribution of individual features and utility-based metrics, the authors argue that the bias intrinsic to their methods has not led to undesirable bias or side-effects in other aspects of the learned distribution. However, the constraints were strict and would be hard to scale. The idea of incorporating knowledge-based constraints in the otherwise naive \gls{gan} is in fact gaining attention (See Section \ref{sec:knowledge} \par

\section{Suggestions of requirements for OHD-GAN development}

\subsection{Models of appropriate scope and equivalent degree of evaluation}\label{sec:basic}
Overall, evaluation methods were superficial or uni-dimensional relative to the scope of the task. As previously discussed, finding convincing and robust evaluation metrics for synthetic health data is an open issue. Weak metrics become a prominent issue when the learning task is broad, loosely defined, constructed for the sole purpose of evaluation, or the scope of application is too large. The difficulty of explaining or validating the realism of data representing a patient, often longitudinal and which factors deferentially contribute to disease characterization makes the assessment of synthetic data ambiguous, thus demanding stronger evidence to claims.\par

\footnotesize
\tcbset{enhanced,before skip=1cm, nobeforeafter, width=0.5\linewidth}
\begin{tcolorbox}[arc=0mm, 
    colback=cadmiumgreen!10!white, 
    coltext=cadmiumgreen!90!black,  
    colframe=cadmiumgreen!90!black,
    colbacktitle=cadmiumgreen!80,
    leftrule=3mm,
    rightrule=0mm, 
    toprule=0mm, 
    bottomrule=0mm, 
    box align=top]

Modelling efforts for OHD-GAN should be limited in scope to develop robust algorithms for a single data type or modality.
\begin{itemize}
    \item This makes qualitative evaluation by visual inspection from experts possible and meaningful.
    \item The behaviour of the model can be assessed straightforwardly
    \item Conditional models are easier to develop.
    \item The evaluation metrics should not be defined solely for the purpose but from a peer-reviewed healthcare publication.
\end{itemize}

\end{tcolorbox}
\hfill
\begin{tcolorbox}[tcbox width=auto, 
    arc=0mm, 
    colback=white, 
    coltext=cadmiumgreen, 
    boxrule=0pt, 
    colframe=white,
    box align=top]
    
\epigraph{\textit{A baby learns to crawl, walk and then run.  We are in the crawling stage when it comes to applying machine learning.}}{\textit{Dave Waters}}

\end{tcolorbox}
\normalsize%
\subsection{Data-driven architecture}\label{sec:archi}

Deep architectures are based on the intuition that multiple layers of nonlinear functions are needed to learn complicated high-level abstractions \cite{Bengio_2009}. \gls{cnn} capture patterns of an image in a hierarchical fashion, such that in sequence, each layer forms a representation of the data at a higher level of abstraction. This type of data-oriented architecture has led to impressive performance for \gls{cnn} and image data.\par

Health data presents a different, analogous multi-level structure. As an illustration, a predictive algorithm developed in a hierarchical structure was shown to form representations of \gls{ehr} that capture the sequential order of visits and co-occurrence of codes within a visit. It led to improved predictor performance, and also allowed for meaningful interpretation of the model \cite{choi2016multi}. Similarly, models of time-series based on a continuous time representation \footnote{Those interested in \gls{gan} for wavelike data will find many examples \cite{Delaney2019,Golany2019,Ye2019,Wang2019d,Singh2020,Aznan2019,Hartmann2018}.}, such as \glspl{eeg} and \glspl{ecg} found in \gls{ehr} data, have shown improved accuracy over discrete time-representations \cite{rubanova2019latent,de2019gru}. Creative adaptations of the data for existing architectures have provided surprising results. For example, \gls{ohd} input into a CNN were transformed to image(bitmaps) in which the pixels encoded the information \cite{Fukae2020}

\footnotesize
\tcbset{enhanced, before skip=1cm, nobeforeafter, width=0.5\linewidth}
\begin{tcolorbox}[
    arc=0mm, 
    colback=cadmiumgreen!10!white, 
    coltext=cadmiumgreen!90!black,  
    colframe=cadmiumgreen!90!black,
    colbacktitle=cadmiumgreen!80,
    leftrule=3mm,
    rightrule=0mm, 
    toprule=0mm, 
    bottomrule=0mm, 
    box align=top]
    
The architecture of \gls{ohd}-GAN should be engineered to match the data, not the other way around. Data with minimal transformations, to the extent possible. In addition to preventing information loss, this ensures models will reflect the real generative process. Such models are more likely to further our understanding about them and the biological drivers. With deeper understanding, novel architecture of higher complexity will be engineered. Furthermore, the learned statistical distribution is inevitably more meaningful and interpretable, facilitating applications in the healthcare domain and supporting the inference of insights. 

\end{tcolorbox}
\hfill
\begin{tcolorbox}[tcbox width=auto, 
    arc=0mm, 
    colback=white, 
    coltext=cadmiumgreen, 
    boxrule=0pt, 
    colframe=white,
    box align=top]

\epigraph{Torture the data, and it will confess to anything.}{\textit{Ronald Coase}}

\end{tcolorbox}
\normalsize%
\subsection{Evolving the patients}
As we have seen, \gls{ohd-gan} are not exclusively used to produce "fake" patients, but also to be representative of an particular patient. Common examples are translating between patient states, or producing counterfactuals. It would be interesting to see if combining \gls{gan} with what is know as evolutionary computing could produced valuable results. We can think of a \gls{gan} transforming the patient data to an alternative state, after which the evolutionary algorithms would optimize this new state in a continuous fashion, as new data about the patient becomes available. Immediately after writing this, a quick search confirms the combination can have impressive results, either in optimizing the evolutionary process \cite{He2020-zm}, exploring the latent space \cite{Schrum2020-vl}, or expanding the information received by the discriminator \cite{Mu2020-id}. 

\footnotesize
\tcbset{enhanced,before skip=1cm, nobeforeafter, width=0.5\linewidth}
\begin{tcolorbox}[arc=0mm, 
    colback=cadmiumgreen!10!white, 
    coltext=cadmiumgreen!90!black,  
    colframe=cadmiumgreen!90!black,
    colbacktitle=cadmiumgreen!80,
    leftrule=3mm,
    rightrule=0mm, 
    toprule=0mm, 
    bottomrule=0mm, 
    box align=top]

Unexpected combinations of existing algorithms can harness the in strengths of both, or compensate for lacking, producing performance above the capabilities of both. Evolutionary algorithms are only one particular example, but in fact we've seen a few in this review, including the first that incorporated an \gls{ae} or the techniques borrowed from the \gls{sdv}. Mix-and-match, select, repeat is the principle behind any \gls{ml} model, the notion of meme, and broadly human knowledge... and our existence.
\end{tcolorbox}
\hfill
\begin{tcolorbox}[tcbox width=auto, 
    arc=0mm, 
    colback=white, 
    coltext=cadmiumgreen, 
    boxrule=0pt, 
    colframe=white,
    box align=top]
    
\epigraph{To me, it is very striking to now understand that their work, described in "ImageNet Classification with deep convolutional neural networks", is the combination of very old concepts (a CNN with pooling and convolution layers, variations on the input data) with several new key insight (very efficient GPU implementation, ReLU neurons, dropout), and that this, precisely this, is what modern deep learning is.}{\textit{Andrey Kurenkov \cite{kurenkov2020briefhistory}}}

\end{tcolorbox}
\normalsize%

\subsection{Forcing, disciplining or guiding \label{sec:knowledge}}

To build statistical models we define rules and relations that they are forced to optimize when learning. On the other hand, \glspl{gan} are given free range in a space of possibilities and are disciplined for exploring certain areas, but are provided no explanation. \par

We build enormous models and let them fight back and forth in a mim-max battle that goes on forever, denying them our valuable knowledge. The idea of introducing human knowledge in the otherwise naive training process has gained some attention.\par

Posterior regularization is usually used to impose constraints on probabilistic models, but \glspl{gan} lack the necessary Bayesian component. In the student-teacher model, where a larger model is used to train a smaller one, the process is knowledge distillation. Such models are developed for many applications, such as compression, improving accuracy and accelerating training \cite{abbasi2019odeling}.

In the field of \gls{rl}, a mathematical correspondence between \gls{ps} and \gls{rl} led to the probabilistic \gls{pr} framework \gls{irl} that seeks to learn a reward function from expert demonstrations. This was followed by approaches capable of learning both the reward function and the policy \cite{finn2016guided,fu2018learning}. \citeauthor{Hu2018} then demonstrated a correspondence between \glspl{rl} and \glspl{gan}. This allowed them to develop a \gls{gan} with a constraint-based learning objective \cite{Hu2018}.\par

The constraints, seen as a reward function, can be learned by the model through an algorithm involving maximum entropy. This means the known constraints can be input directly or partially and left to be learned automatically. The algorithm consistently improved the speed and quality of training, and accuracy on a few tasks. The approach is exemplified on an image translation task where images of people are transformed from one pose (ex. looking forward) to another (ex. head turned left). The constraint is provided by a pre-trained auxiliary classifier that assigns each pixel to a body part, and is adapted jointly with the \gls{gan}. The \gls{gan} is rewarded for preserving the mapping in the output image. A performance comparison against unconstrained and fixed-constraint models results in similar training loss and evaluation metric. However when evaluated by humans, the novel approach surpasses the other models on 77\% of test cases. \par

\footnotesize
\tcbset{enhanced, before skip=1cm, nobeforeafter, width=0.5\linewidth}
\begin{tcolorbox}[
    arc=0mm, 
    colback=cadmiumgreen!10!white, 
    coltext=cadmiumgreen!90!black,  
    colframe=cadmiumgreen!90!black,
    colbacktitle=cadmiumgreen!80,
    leftrule=3mm,
    rightrule=0mm, 
    toprule=0mm, 
    bottomrule=0mm, 
    box align=top]
    
The prospect of GANs being able go incorporate auxiliary information or constraints they  can automatically learn to optimize is  a golden research opportunity. This would empower them with the prior knowledge until now reserved to model in the category of the same name, while keeping their  ability to learn in an unsupervised adversarial framework.

\end{tcolorbox}
\hfill
\begin{tcolorbox}[tcbox width=auto, 
    arc=0mm, 
    colback=white, 
    coltext=cadmiumgreen, 
    boxrule=0pt, 
    colframe=white,
    box align=top]

\epigraph{“If you had all the world’s information directly attached to your brain, or an artificial brain that was smarter than your brain, you’d be better off.”}{\textit{Sergey Brin}}

\end{tcolorbox}
\normalsize%

\subsection{Interpretability\label{sec:latent-space}}
Even though a few authors attempted to understand the behavior of their models, overall the subject was left largely unmentioned. It is imperative that future experimentation and publication give equal importance to the interpretation of their models and establishing means to do so. In the healthcare domain, black box machine learning models find little adoption, and synthetic data is most often met with dismissal to its validity. The task is not impossible, as for any other opaque system, and in fact experimental sciences in general. The simplest approach is to  provide input, observe the output, reformulate our hypotheses, and modify the input accordingly. Repeatedly, to convergence. Fortunately, in this case the internals workings are entirely available, tipping the balance between brute-force, and knowledgeable-driven exploration of the system. In addition, we believe "qualitative" evaluation by visual inspection has much greater potential, still to be defined. What better to define interpretation than a medical professional decoding the hidden relations in data visually. \par

In theory, the latent space is a lower-dimensional representation of basic concepts that should be directly interpretable. However, in practice these concepts are entangled over multiple nodes. In what is a preliminary, but encouraging proof-of-concept, \cite{lui2019-latent} explore how they can use perturbations to reveal patterns in a \gls{beta-vae} trained to capture brain structure in mice. By generating a collection of images from a dense interpolation of the latent space, they were able to examine the projective field of latent variables onto the pixels. They found zones of high variance that corresponded to biologically relevant areas. Reversing the experiment, they masked areas of the images and found that many latent factors were not activated by all regions of interest and had localized receptive fields. Whereas complex highly connected regions such as the hippocampus activated almost all latent factors. Curiously, the projective and receptive fields may not be aligned. Numerous other publications have shown that they capture meaningful properties and structure of the data, reducing complexity to a level that lends itself to interpretation \cite{Way2020, Koumakis2020}. In one instance involving transcription factor micro-array data, a close one-to-one mapping could be obtained from the last hidden layer, in addition to the higher level layers that related to biological processes in a hierarchical fashion \cite{chen2016-latentyeast}. Pushing the boundaries further, by correlating the output features of a GAN with the latent space dimensions allowed controllable semantic manipulation of the generated data \cite{Wang2020latent,Ding2020latent,Li2020latent}. However, a recent information-theoretic \gls{gan} simplified interpretation greatly by forcing the latent nodes to learn disentangled representations. In addition to adversarial loss, \gls{info-gan} also maximizes the mutual information between small numbers of latent nodes. The result is highly interpretable nodes that represent distinct concepts that can be easily influenced, or in some cases interpolate smoothly between features \cite{Chen2016c}.\par

\subsection{Benchmarking, a priority}
It became slowly obvious through the secession of experiments that there is a glaring problem of standardization of evaluation. New algorithms and applications are being demonstrated at an increasing rate. On the contrary, standardized benchmarks, procedures to transform the data, and source has remained scarce, one can hardly compare the models objectively or nominate the best performances. Commendably, \citeauthor{Camino2020bench} are the first to bring attention to this issue in a position paper that provides quantitative arguments. Notably the myriad of ways commonly used datasets are reprocessed, metrics that are not comparable, and hyperparater sweep results, for which no transformation code and optimal values are released and the lack of effort towards reproducibility will only reduce credibility of the field. On a positive note, we've compiled a list of the repositories which were made open-source in Table \ref{tab:5:sourcecode} and a list of the common dataset links can be found in Table \ref{tab:5:sourcecode}.\par
In this regard the replication of medical studies with synthetic data by \citeauthor{Yale_2020} substantiate the value of \gls{sd} for exploratory data analysis, reproducibility on restricted data and more generally education in scientific training \cite{Reiner_Benaim2020-lx}. Reproducing medical or clinical studies will be necessary to gain mainstream adoption of \gls{gan} produced \gls{sd} and dispel the scepticism it is generally met with. The medical domain is known for its slow pace in adopting new technologies and predictive \gls{ml} is still far from meeting its full implementation potential \cite{Qayyum2020-ir}. Medical professionals care foremost about the well-being of their patients and will only consider results obtained from synthetic data if they have the assurance that they are valid \cite{Rankin2020}.  A remarkable resource for the purpose of benchmarking is the clinical prediction benchmarks defined on the \gls{mimic} data by \citeauthor{harutyunyan_multitask_2019}. The tasks are clearly defined and the source code to process the data and the algorithms is available \cite{harutyunyan_multitask_2019}. We suggest comparing the accuracy of the predictive algorithms applied to the original data versus the synthetic data to be evaluated. However, concerted efforts and informal guidelines that can be agreed upon should be on a regular schedule. We fully support the idea or organized challenges and hackahton proposed by \cite{Camino2020bench} and suggest a progressive approach to realizing it.\par

\subsubsection{Ultra-open source, collaborative, publishing communities}
In a successful and educative experiment on collaborative writing and crowd-sourcing, an article was entirely written in an open-source GitHub repository. Anyone willing to add their knowledge to the publication was welcome to do so, reaching 30+ authors in 20 countries. Every change proposal is requested for inclusion by a Pull Request, for which R2-3 approvals are necessary. Withing minutes, automated deployment procedures (Github since then released Actions, requiring minimal coding), took care of verifying compliance to guidelines, citation management, DOI registration, and compilation of latex or Markdown. Withing minutes a revised  document is released, making the publication a contiguously up-to-date source of knowledge, that can be augmented in the web version with interactive code-books and figures.\par
Issues can be discussed in the appropriate channels, but most importantly the nature of GitHub ensures attribution of work done, down to a single character. The authors also implemented immutable backup on the blockchain. Since then distributed storage and computation blockchains have reached maturity and could store models, training artefacts, and data for competition at a trivial cost. As an alternative,  the \href{http://bit.ly/WandB-ML}{Weights and Biases (WandB)} platform is a fitting environment, worth a look even for individuals. The traditional publishers have long been touting a makeover of the publication system, changes are slow and trivial, whereas den centralized, person to person, systems have been transforming whole sectors faster than ever.\\

\section{Directions for future research}
\subsection{Building a patient model}
The ultimate goal for generative models of \gls{ohd} must be to develop an algorithm capable of learning an all encompassing patient model. It would then be possible to generate full \gls{ehr} records on demand, integrating genetic, lifestyle, environmental, biochemical, imaging, clinical information into high-resolution patient profiles \cite{Capobianco2020}. This is in fact the intention of the patient simulator Synthea. However, Synthea will eventually face a problem with scalability and the capacity of semi-independent state-transition models to coordinate in capturing long-range correlations.\par

Once basic models of health data, as described in Section \ref{sec:basic}, have been developed and validated, these can be progressively combined in a modular fashion to obtain increasingly complex patient simulators. Furthermore, having designed the architecture of these basic models on the underlying data in a way that is comprehensible, as described in \ref{sec:archi}, will facilitate the composition of more complex models. Inputs, outputs and parts of these models can be conditionally attached to others such that the generative process occurs in a way that reflects the real generative process.

\subsection{Evaluating complex patient models \label{sec:evaluation-cqm}}
Once more complex models are developed, the problem is again finding meaningful evaluation metrics of data realism. Capobiano et al. insist on the necessity for data performance metrics encompassing diagnostic accuracy, early intervention, targeted treatment and drug efficacy \cite{Capobianco2020}. In their publication exploring the validation of the data produced by Synthea, Chen et al. provide an interesting idea to achieve this \cite{Chen_2019}. Noting that the quality of care is the prime objective of a functional healthcare system, they suggest using \glspl{cqm} to evaluate the synthetic data. These measures "are evidence-based metrics to quantify the processes and outcomes of healthcare", such as "the level of effectiveness, safety and timeliness of the services that a healthcare provider or organization offers."(Chen 2019). High-level indicators such as \glspl{cqm} domain specific measures of quality, are specifically designed for higher level or multi-modal representations of healthcare data. The constraints introduced in \gls{heterogan} should be leverage to evaluate the realism of the synthetic data, rather than bias the generator training. Composing a comprehensive set of such constraints could possibly serve as a standardized benchmark.
At the individual level, Walsh et al. employ domain specific indicators of disease progression and worsening and compare agreement of the simulated patient trajectories with the factual timelines \cite{walsh2020generating}.\par
In addition to \gls{cqm}, we propose the use of the Care maps used by the Synthea model to simulate patient trajectories as evaluation metrics \cite{Walonoski_2017}. Care maps are transition graphs developed from clinician input and Clinical Practice Guidelines, of which the transition probabilities are gathered from health incidence statistics. While these allow the Synthea algorithm to simulate patient profile with realistic structure, they also prevent it from reproducing real-world variability. Conversely, while \glspl{gan} have the ability to reproduce the quirks of real data, they also lack the constraints preventing nonsensical outputs. As such, Care maps provide an ideal metric to check if the synthetic data conforms to medical processes.\par 
In fact, this has been used before in a competition where participants were given synthetic data from finite state transition machines with known probabilities and tasked to build and learn models that would reproduce those of the original, unseen models. The participants worked according to the Perplexity metric, commonly used in NLP, which quantifies how well a probability distribution or probability model predicts a sample \cite{Verwer_2013}. We postulate that the Synthea models built with real-world probabilities would provide a unique and robust way to evaluate synthetic data according to the metric proposed above, among other means to utilize the state-transition in Synthea and their modularity.

\subsubsection{Opportunities and application to current events}
Synthetic and external controls in clinical trials are becoming increasingly popular \cite{Thorlund2020}. Synthetic controls refer to cohorts that have been composed from real observational cohorts or \gls{ehr} using statistical methodologies. While the individuals included in the cohorts are usually left unchanged, micro-simulations of disease progression at the patient level are used to explore long-term outcomes and help in the estimation of treatment effects \cite{Thorlund2020, Etzioni2002}. Synthetic data generated by \glspl{gan} could be transformative for the problem of finding control cohorts.\par
With the COVID-19 pandemic scientists have become increasingly aware of and vocal about the need for data sharing between political borders \cite{Cosgriff_2020,Becker_2020,McLennan_2020}. An obvious application is generating additional amounts of data in the early stages of the pandemic, potentially creating opportunities earlier. Synthetic data is not only an opportunity to facilitate the exchange of data, but also to adjust the biases of samples obtained from different localities. Factors such as local hospital practices, different patient populations and equipment introduce feature and distribution mismatches \cite{Ghassemi2020}. These disparities can be mitigated by translation of \gls{gan} algorithms, such as \gls{cycle-gan} proposed by Yoon et al.

    \section{Source-code and datasets}
The algorithms presented in this review can undoubtedly find usefulness for other health data or similar problems. Most importantly they can be reevaluated on other datasets or improved by adapting them with latest ML techniques. We present in Table \ref{tab:5:sourcecode} a list of links to the source code published by the authors. In addition, we present in Table \ref{tab:5:datasets} the datasets which were employed by the authors in their experiments, for those who were referenced and available. A broad variety of articles about generative and predictive algorithms published along with the source-code can be on \href{https://paperswithcode.com}{Papers With Code} in the \href{https://paperswithcode.com/area/medical}{medical section}. Notably, they host a yearly ML Reproducibility Challenge to "[...] encourage the publishing and sharing of scientific results that are reliable and reproducible." in which papers accepted for publication in top conferences are evaluated by members of the community reproducing their experiments \cite{Sinha}. Benchmarks are also presented on the website, but unfortunately \gls{corgan} is the only entry in the medical section. 

\begin{table}[H]
    \footnotesize
    \setlength{\extrarowheight}{0.15em}
    \caption{Open-source repositories \label{tab:5:sourcecode}}
    
    \begin{tabularx}{\linewidth}{@{}XXp{3cm}p{1cm}@{}}\toprule
        Author and algorithm & Repository & Format & Data\\ \midrule
        
        \citeauthor{baowaly_2019_IEEE} \gls{medbgan} 
        & \href{https://github.com/baowaly/SynthEHR}{baowaly/SynthEHR}
        & Tensorflow 
        & \checkmark \\
        
        \citeauthor{baowaly_2019_jamia} \gls{medbgan}, \gls{medwgan} 
        & \href{https://github.com/baowaly/SynthEHR}{baowaly/SynthEHR} 
        & Tensorflow 
        & \checkmark \\
        
        \citeauthor{severo2019ward2icu} \gls{cwgan-gp} 
        & \href{https://github.com/3778/Ward2ICU}{3778/Ward2ICU} 
        & PyTorch
        & \ding{54}\\

        \citeauthor{torfi2019generating} \gls{corgan}
        & \href{https://github.com/astorfi/cor-gan}{astorfi/cor-gan} 
        & PyTorch
        & \checkmark \\
        
        \citeauthor{Jackson_2019} \gls{medgan}
        & \href{https://github.com/marcolussetti/extended-medgan}{marcolussetti/extended-medgan} 
        & Tensorflow
        & \checkmark\\
  
        \citeauthor{Beaulieu-Jones2019-ct} \gls{ac-gan} 
        & \href{https://github.com/greenelab/SPRINT_gan}{greenelab/SPRINT\_gan} 
        & Keras 
        &\checkmark \\
        
        \citeauthor{Xu2019-ay} \gls{ctgan}
        & \href{https://github.com/sdv-dev/TGAN}{sdv-dev/TGAN} 
        & Tensorflow
        & \checkmark \\
        
        \citeauthor{yale2019ESANN} \gls{healthgan}
        & \href{https://github.com/yknot/ESANN20193}{yknot/ESANN2019} \href{https://competitions.codalab.org/competitions/19365}{Codalab 19365} 
        & Tensorflow
        & \checkmark\\
        
        \citeauthor{Yale_2020} \gls{healthgan}
        & \href{https://github.com/TheRensselaerIDEA/synthetic_data}{TheRensselaerIDEA/synthetic\_data} 
        & Tensorflow, 
        & \checkmark\\
        
        \citeauthor{tanti2019} \gls{dp-auto-gan}
        & \href{https://github.com/DPautoGAN/DPautoGAN}{DPautoGAN/DPautoGAN}
        & PyTorch
        & \checkmark\\
        
        \citeauthor{BaeAnomiGAN2020} \gls{anomigan}
        &\href{https://github.com/hobae/anomigan}{hobae/anomigan} 
        & Tensorflow, Keras
        & \checkmark\\
        
        \citeauthor{zhu_2020} \gls{glugan}
        & \href{https://bitbucket.org/deep-learning-healthcare/glugan}{deep-learning-healthcare/glugan}
        & Tensorflow
        & - \\
        
        \citeauthor{chen2019ganleaks} \gls{medgan}, \gls{wgan-gp}, DC-GAN
        & \href{https://github.com/DingfanChen/GAN-Leaks}{DingfanChen/GAN-Leaks} 
        & PyTorch
        & \checkmark\\
        
        \midrule 
        \multicolumn{4}{c}{Source code was linked in the publication and could not be found.} \\ 
        \midrule 
        \multicolumn{4}{p{\textwidth}}{      
        \citeauthor{chin2019generation} , \citeauthor{Jordon2019} \gls{pate-gan}, \citeauthor{chu2019treatment} \gls{adtep},\citeauthor{yu2019rare} \gls{ssl-gan}, \citeauthor{Yang_2019_cdss} \gls{cgan}, \citeauthor{Yang_2019_ehr} \gls{gcgan}, \citeauthor{Yang_2019_impute_ehr} \gls{cgain}, \citeauthor{walsh2020generating} Adversarial \gls{crmb}, \citeauthor{Fisher2019} Adversarial \gls{crmb} \citeauthor{cui2019conan} \gls{conan},\citeauthor{chincheong2020generation} \gls{wgan-dp},\citeauthor{Zhang2020} \gls{emr-wgan},\citeauthor{yan2020generating} \gls{heterogan},\citeauthor{ozyigit2020generation} \gls{rsdgm}, \citeauthor{Yoon2020-anon} \gls{ads-gan}, \citeauthor{Goncalves2020} \gls{mc-medgan}} \\
        
        \bottomrule
    \end{tabularx}
\end{table}

\begin{table}[H]
    \footnotesize
    \setlength{\extrarowheight}{0.35em}
    \caption{Relevant dataset used in the publications\label{tab:5:datasets}}
    \begin{tabularx}{\textwidth}{@{}Xp{5cm}@{}}\toprule
    Dataset & Link\\\midrule
    
    SPRINT Clinical Trial Data \cite{wright2016randomized} 
    & \href{https://challenge.nejm.org/pages/home}{SPRINT Data Analysis Challenge}\\
    
    Coalition Against Major diseases Online Repository for AD \cite{Neville_2015} 
    &\href{https://c-path.org/programs/dcc/projects/alzheimers-disease/coalition-against-major-diseases-consortium-database-camd-admci/}{CAMD AD/MCI}\\
     
    American Time Use Survey (ATUS) \cite{us_bureau_of_labor_statistics_american_nodate}
    & \href{https://www.bls.gov/tus/}{ATUS}\\

    Philips eICU \cite{pollard2018eicu}
    & \href{https://physionet.org}{Physionet \cite{Goldberger_2000}}\\
    
    Multiparameter Intelligent Monitoring in Intensive Care (MIMIC-III v1.4) \cite{Johnson_2016}
    & \href{https://mimic.physionet.org}{Physionet} \cite{Goldberger_2000}\\
    
    Vanderbilt University Medical Center Synthetic Derivative \cite{Roden_2008}   
    & \href{https://victr.vumc.org/biovu-description/}{BioVU}\\
    
    UC Irvine Machine Learning Repository \cite{Dua:2019}  
    & \href{http://archive.ics.uci.edu/ml/index.php }{UCI ML repository}\\
    
    Ward2ICU \cite{severo2019ward2icu} 
    & \href{https://arxiv.org/abs/1910.00752}{ArXiv}\\
    
    SEER Cancer Statistics Review (CSR) \cite{noone2018cronin} 
    & \href{https://seer.cancer.gov/data/access.html}{SEEr Incidence database}\\
    
    PREAGRANT \cite{Fasching_2015} & 
    On request: \href{mailto:peter.fasching@uk-erlangen.de}{peter.fasching@uk-erlangen.de} \\
    
    New Zealand National Minimum Dataset (hospital events) \cite{events}
    & \href{https://www.health.govt.nz/nz-health-statistics/access-and-use/data-request-form}{Data request form}\\
    
    Sutter Palo Alto Medical Foundation (PAMF)  Heart failure study \cite{Choi2017-nt} 
    & \href{https://www.sutterhealth.org/research/pamfri}{PAMFRI}\\
    
    \bottomrule
    \end{tabularx}
\end{table}
    \section{Conclusion}
\normalsize
\Gls{sd} has been a subject of interest for quite some time, with officials seeing enough value to launch longitudinal state-wide endeavours such as the Synthetic Data Project (SDP), funded by the United States Department of Education(USDOE) \cite{Bonnery2019-ug}. They dismiss a series of anonymization techniques, stating the burden on worker and financial resources, and the privacy guarantees that would not sufficient for governmental agencies. Issues that have only gained weight with the accumulation of big data, and the number of new sources growing consistently. The questions they hoped to answer at the start of the project in 2016 are still not fully answered (evaluation, scientific validity, legal implications). Their 2019 report on the experience is packed with interesting insights. Noting the distrust people tend to have of synthetic data, they were the ones who first proposed the idea conducting experiments on synthetic data, that could then be confirmed on real data by simply sending the analysis to the data holder (with the logistics described extensively and augmented by a flowchart).\par

The publication ends with a series case reports. The instances where the data could not satisfy requirements are analyzed with the aim of informing similar projects in the future. However the bulk of reports describe cases where was highly applicable. They concluded by predicting that the cost of generating \gls{sd} will diminish and that the methods to do so will improve. Their hopes for \gls{sd} include: easier access for researcher to the wealth of data, increased access providing downstream benefits at the state level, the these benefits encourage others to undertake similar projects that would increase generalizability of findings across states, and a preference for open data.

\renewcommand{\epigraphsize}{\footnotesize}
\setlength{\epigraphwidth}{12cm}
\epigraph{
    "[although some argue for] having secured data centers for administrative data utilization [...], our experience suggests that such centers may not solve the desire for fast turn-around research or broaden access to those with unique perspectives. Synthetic data represent a promising approach for increasing easy access to secure data while simultaneously protecting the confidentiality of individuals."}{\textit{Daniel Bonnéry, Yi Feng, Angela K. Henneberger, Tessa L. Johnson,\\ Mark Lachowicz, Bess A. Rose, Terry Shaw,\\ Laura M. Stapleton, Michael E. Woolley and Yating Zheng}}
    
The \gls{gan} was devised in 2014 in Montreal, Canada by \citeauthor{goodgan} at Université de Montreal. Two years before the start of the SDP, which must of been planned over a few years. It was too early for them to know about this obscure technique based on two neural networks competing against each other. Since then, \href{http://papers.nips.cc/paper/5423-generative-adversarial-nets}{Generative Adversarial Networks} \cite{goodgan} has inspired \textbf{23805} citations and algorithms capable of synthesizing data of impeccable similitude. We have surveyed a multitude of \gls{gan} algorithms built on the same basic idea of trial and error against an opponent that learns your faults. Despite the simple concept, we've seen that their range of application is wide, in general as well as in the healthcare domain. The variety of architectures and techniques we've seen reflect the heterogeneity of health data. Seemingly the difficulty of achieving stable learning with \glspl{gan} in general delayed their application to \gls{ohd}, while in medical imaging the development boomed much earlier sustained by the success of \gls{cnn} in other fields. Notably, the innovation in the field of \gls{cnn} has not slowed down after a few algorithms obtained excellent performance in image classification, but has deepened and branched out. Research concerning \gls{ohd} seems to be gaining momentum rapidly. We saw thoughtfully engineered algorithms designed for the characteristics of \gls{ohd}. Crucially however, pushing the research further will require a community effort to discover and defined metrics and standards upon which we can base objective assessment of models and \gls{sd}. The challenges posed by \gls{ohd} are nothing but encouragement for investigation and interpretation that can further our understanding of \glspl{gan}, machine learning and human health. Undoubtedly the innovations made for \gls{ohd} will find matches in other fields which may share the same data troubles. 
\pagebreak

    \pagebreak

    \printglossary[type=oalgo]
    \printglossary[type=\acronymtype]

    \pagebreak

    \bibliography{biblio}

\end{document}